\documentclass[10pt,twocolumn,letterpaper]{article}
\usepackage[pagenumbers]{cvpr}

\usepackage{times}
\usepackage{epsfig}
\usepackage{graphicx}
\usepackage{amsmath,amssymb,amsfonts,amsthm}
\usepackage{bm}
\usepackage{booktabs}
\usepackage{multirow}
\usepackage{xcolor}
\usepackage{array}
\usepackage{tabularx}
\usepackage{enumitem}
\usepackage{xspace}
\usepackage[ruled,vlined]{algorithm2e}
\SetKwInput{KwInput}{Input}
\SetKwInput{KwOutput}{Output}

\newtheorem{proposition}{Proposition}
\newtheorem{theorem}{Theorem}
\newtheorem{lemma}{Lemma}

\newcommand{\method}{BISECT\xspace}
\newcommand{\R}{\mathbb{R}}
\newcommand{\E}{\mathbb{E}}
\newcommand{\norm}[1]{\left\|#1\right\|}

\newcommand{\trace}{\mathrm{tr}}
\newcommand{\diag}{\mathrm{diag}}
\newcommand{\spanop}{\mathrm{span}}
\newcommand{\rsvd}{\mathrm{RSVD}}
\newcommand{\qr}{\mathrm{QR}}
\newcommand{\gain}{\mathcal{R}}
\newcommand{\Gpast}{G_{<t}}
\newcommand{\Gcur}{G_{t}}
\newcommand{\Gtot}{G_{1:t}}
\newcommand{\Gwhit}{\widehat{G}_{t}}
\newcommand{\Pinv}{P_{\mathrm{inv}}}
\newcommand{\Pnov}{P_{\mathrm{nov}}}
\newcommand{\Qinv}{Q_{\mathrm{inv}}}
\newcommand{\Qnov}{Q_{\mathrm{nov}}}
\newcommand{\OmInv}{\Omega_{\mathrm{inv}}}
\newcommand{\kInv}{k_{\mathrm{inv}}}
\newcommand{\kNov}{k_{\mathrm{nov}}}
\newcommand{\AFinal}{A_{\mathrm{Final}}}
\newcommand{\AMean}{\bar{A}}

\definecolor{cvprblue}{rgb}{0.21,0.49,0.74}
\usepackage[pagebackref,breaklinks,colorlinks,allcolors=cvprblue]{hyperref}

\def\confName{CVPR}
\def\confYear{2026}
\def\paperID{0000}

\title{\method{}: Bilevel Spectral Decomposition for Low-Rank Continual Learning}

\author{
Tom Saliencro\\
University of California, Irvine
\and
Wen-Hsuan Liu\\
University of California, Irvine
\and
Rohan Desai\\
University of Washington
\and
Amara Okonkwo\\
University of Washington
\and
Maya Lindqvist\\
University of California, Irvine
}

\begin{document}
\maketitle

\begin{abstract}
Exemplar-free continual learning on pre-trained Vision Transformers increasingly relies on Low-Rank Adaptation (LoRA), yet prevailing methods govern the low-rank update space through strict subspace isolation. By projecting each incoming task into the approximate null space of historical token activations, isolation-based learners equate any cross-task feature overlap with catastrophic interference. We show that this orthogonality premise induces a structural bottleneck we term \emph{spectral starvation}: when down-projection bases are locked before task optimization, the gradient norm reaching the trainable up-projection is upper-bounded by the energy of current-task features captured by the chosen basis. Because natural visual tasks share low- and mid-level covariance structure, historical null tails contain barely $5.5\%$ of incoming task variance, starving new-class plasticity while blocking positive consolidation along shared directions. We propose \method{} (\textbf{Bi}level \textbf{S}p\textbf{ect}ral Decomposition), a continual LoRA framework that partitions a fixed rank-$r$ budget into two orthogonal functional blocks: an \emph{invariant carrier} spanned by the dominant modes of the joint historical-and-current Gram matrix, and a \emph{novelty complement} extracted via generalized Rayleigh whitening followed by orthogonal complement projection. After single-task training, \method{} folds both blocks back into the frozen backbone using a closed-form precision gate that scales each rank-1 invariant atom by its current-to-joint Rayleigh ratio while preserving novelty atoms at unit gain. Across eight incremental settings on ImageNet-R, ImageNet-A, CIFAR-100, CUB-200, and DomainNet, \method{} outperforms prompt, adapter, and LoRA baselines with zero extra inference parameters or FLOPs, and remains complementary to sparse analytic expansion.
\end{abstract}

\section{Introduction}
\label{sec:intro}

Class-incremental learning (CIL) seeks to equip a visual recognition system with new categories over sequential sessions without storing raw exemplars or querying task identifiers at test time \cite{zhou2024continual}. With pre-trained Vision Transformers (ViTs) now serving as the standard representation backbone \cite{dosovitskiy2021vit}, the locus of continual adaptation has shifted from full-network weight regularization to lightweight parameter-efficient modules, most notably Low-Rank Adaptation (LoRA) \cite{hu2022lora}. The same stability--plasticity tension appears in compact biological circuits: the fly olfactory pathway expands a shared code and then writes plasticity only on a sparse subset of that code \cite{zou2025structural}, and FlyLoRA imports the rank-wise version of that idea into a single LoRA module \cite{zou2025flylora}. Because a LoRA module constrains the weight increment $\Delta W = Q P$ to a rank-$r$ bottleneck ($r \ll \min(n,m)$), and because fixing the down-projection $P \in \R^{r\times m}$ confines all subsequent gradient updates to $\spanop(P^\top)$, the central algorithmic question in LoRA-based CIL is how to construct and consolidate those $r$ basis directions across tasks.

Mainstream LoRA continual learners answer this question through \emph{subspace isolation}. Methods such as InfLoRA \cite{liang2024inflora}, O-LoRA \cite{wang2023olora}, and null-space prompt tuning \cite{lu2024vptnsp,wang2021adamnscl} restrict $P$ to the empirical null space of token features accumulated over prior sessions, while BiLoRA \cite{zhu2025bilora} and PLAN \cite{wang2025plan} enforce near-orthogonality across task branches. The underlying assumption is appealingly simple: any feature direction shared between past and present tasks is a conduit for overwriting old knowledge and must be excluded.

\begin{figure*}[t]
    \centering
    \includegraphics[width=0.98\textwidth]{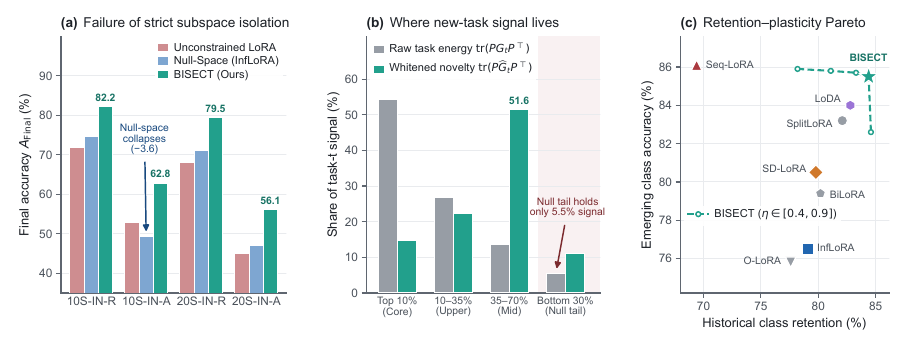}
    \caption{Diagnosing the orthogonality trap in LoRA continual learning. \textbf{(a)} Final accuracy $\AFinal$ across four challenging splits: unconstrained LoRA overwrites prior sessions, whereas null-space projection (InfLoRA) collapses under severe distribution shift on 10S-ImageNet-A. \textbf{(b)} Distribution of task-$t$ signal across quartiles of the historical eigenspectrum $\Gpast$ on ViT-B/16: the bottom $30\%$ null tail holds only $5.5\%$ of raw task energy, starving gradient flow, whereas generalized Rayleigh whitening shifts selection to the mid-spectrum where unexplained variance peaks ($51.6\%$). \textbf{(c)} Historical retention vs.\ emerging-class accuracy on 10S-ImageNet-R; varying the spectral threshold $\eta$ traces a Pareto frontier above isolation and gradient-splitting baselines.}
    \label{fig:motivation}
\end{figure*}

Yet empirical evidence across diverse visual streams reveals that strict isolation trades catastrophic forgetting for a severe loss of plasticity. As shown in Figure~\ref{fig:motivation}(a), while null-space projection (InfLoRA) improves over unconstrained sequential LoRA on rendition shifts (ImageNet-R), it drops below unconstrained LoRA on natural adversarial examples (10S-ImageNet-A, $49.35\%$ vs.\ $52.96\%$). Figure~\ref{fig:motivation}(b) exposes the geometric root cause. Partitioning the eigenspectrum of the historical Gram matrix $\Gpast$ into quartiles shows that more than $80\%$ of a new visual task's feature energy resides in the top $35\%$ of historical directions, whereas the bottom $30\%$ null tail captures only $5.5\%$. Forcing $P$ into that tail starves the up-projection $Q$ of backpropagated gradient energy (\emph{spectral starvation}) while simultaneously forbidding the model from refining shared visual primitives in the leading core. Conversely, choosing directions purely by raw current-task energy $\trace(P\Gcur P^\top)$ rediscovers the historical core and triggers severe drift. What matters for task-specific plasticity is not inactivity in the past, nor raw variance in the present, but the \emph{ratio} of current variance to historical variance—which peaks in the mid-spectrum ($51.6\%$ of whitened novelty in Figure~\ref{fig:motivation}(b)).

Motivated by this diagnosis, we propose \method{} (\textbf{Bi}level \textbf{S}p\textbf{ect}ral Decomposition for Low-Rank Continual Learning). Rather than treating continual adaptation as a homogeneous isolation problem, \method{} splits a stationary rank-$r$ budget into two orthogonal subspaces with distinct functional roles and distinct post-task consolidation rules:
\begin{enumerate}[leftmargin=1.4em]
    \item \textbf{Invariant Carrier ($\Pinv \in \R^{\kInv\times m}$)}: Extracted from the dominant $\kInv$ singular vectors of the joint Gram matrix $\Gtot = \Gpast + \Gcur$ via randomized subspace iteration, capturing cross-task shared structure.
    \item \textbf{Novelty Complement ($\Pnov \in \R^{\kNov\times m}$)}: Constructed by Cholesky whitening $\Gcur$ against the damped historical Gram $\Gpast + \rho I$ to maximize the generalized Rayleigh quotient, followed by exact projection onto $\spanop(\Pinv^\top)^\perp$ and QR orthonormalization.
    \item \textbf{Precision-Weighted Consolidation ($\OmInv^\star$)}: After optimizing $Q = [\Qinv, \Qnov]$ on task $t$, \method{} folds the adapter into the frozen backbone by scaling each rank-1 invariant atom by its closed-form posterior precision ratio $\omega_j^\star \in [0,1]$, while folding novelty atoms at unit weight.
\end{enumerate}

Our contributions are fourfold:
\begin{itemize}[leftmargin=1.3em]
    \item We formalize \emph{spectral starvation} in frozen-basis LoRA continual learning (Proposition~\ref{prop:starvation}), proving that null-space isolation bounds the up-projection gradient norm by the residual tail energy of incoming features.
    \item We introduce \method{}, a bilevel spectral decomposition that pairs a joint-Gram invariant carrier with a Cholesky-whitened novelty complement and allocates rank layer-adaptively across transformer blocks.
    \item We prove that our novelty basis solves a constrained generalized Rayleigh problem (Theorem~\ref{thm:sir}) and derive a closed-form rank-1 consolidation gate that strictly minimizes cumulative representation drift (Theorem~\ref{thm:recal}).
    \item Across eight class-incremental benchmarks on a frozen ViT-B/16 protocol, \method{} leads prompt and LoRA baselines while reducing subspace setup latency by $12.4\times$ relative to LoDA and adding zero inference overhead. Sparse analytic expansion \cite{zou2026flycl} remains stronger on CIFAR-100 and CUB, where a frozen code is already linearly separable.
\end{itemize}

\section{Related Work}
\label{sec:related}

\paragraph{Pre-trained continual learning with prompts and adapters.}
Classical continual learning relies on episodic replay, parameter importance regularization, or knowledge distillation \cite{zhou2024continual}. With strong pre-trained ViTs \cite{dosovitskiy2021vit}, rehearsal-free methods instead attach compact tunable modules to a frozen encoder. Prompt-tuning approaches such as L2P \cite{wang2022l2p}, DualPrompt \cite{wang2022dualprompt}, CODA-Prompt \cite{smith2023codaprompt}, CPG \cite{gao2024cpg}, and VQP \cite{jiao2024vqp} maintain key-value prompt pools, while recent extensions introduce cross-modal prompt generators \cite{hu2026drape}. Parallel adapter frameworks adapt transformer blocks via semantic shift compensation (SSIAT \cite{tan2024ssiat}), expandable subspace ensembles (EASE \cite{zhou2024ease}, Miles \cite{jiang2026miles}), dynamic adapter integration \cite{li2025dynamic}, model surgery (MOS \cite{sun2025mos}), self-expanding adapter mixtures \cite{wang2025self}, cross-subspace alignment (CKAA \cite{he2025ckaa}), elastic knowledge preservation (EKPC \cite{wang2026ekpc}), or task-conditioned transformations \cite{fu2026oneadapter}. While effective, expanding prompt pools or adapter ensembles increases inference latency or parameter count with task horizon $T$, and does not address how to partition directions inside a single merged weight update.

\paragraph{Low-rank subspace governance in continual learning.}
LoRA \cite{hu2022lora,oh2026sdslora} enables zero-inference-overhead continual learning because learned rank-$r$ updates can be folded directly into the backbone weights after each session. To mitigate forgetting, O-LoRA \cite{wang2023olora}, InfLoRA \cite{liang2024inflora}, and VPT-NSP \cite{lu2024vptnsp} project new down-projections into the approximate null space of past feature covariances, extending classical null-space SGD \cite{wang2021adamnscl}. BiLoRA \cite{zhu2025bilora} and PLAN \cite{wang2025plan} enforce pairwise orthogonality via bilinear parameterization or proactive rank allocation, while SD-LoRA \cite{wu2025sdlora} decouples directional bases from learnable magnitudes. More recent works relax strict isolation: CoSO \cite{cheng2025coso} continuously updates subspace bases, LoRA-P\&M \cite{qiu2025lorapm} combines perturbation training with post-hoc merging, SplitLoRA \cite{qiu2025splitlora} splits optimization gradients into primary and secondary spaces, and LoDA \cite{he2026loda} selects bases via raw projection energy alongside an auxiliary router. Related LoRA architectures explore balanced updates (Janus-LoRA \cite{chen2026januslora}), layer-scaled residual gradients (KeepLoRA++ \cite{luo2026keeplorapp}), geometric subspace evolution (Geo-LoRA \cite{feng2026geolora}), implicit rank-wise mixture-of-experts (FlyLoRA \cite{zou2025flylora}), or nested expansion for multimodal instruction tuning \cite{che2025lora,wang2025harmonious,liu2026acelora}. FlyLoRA keeps one adapter and treats ranks as experts, but its top-$k$ mask also zeros the forward pass, so the test-time rank is strictly smaller than $r$ \cite{zou2025flylora}.

\method{} does not banish historically active directions: an invariant carrier stores shared primitives, and a precision gate scales their write-back. Unlike LoDA, which keeps $2.36$M extra routing parameters at test time, SplitLoRA, which partitions gradients after the basis is fixed, and FlyLoRA, which drops ranks at inference \cite{zou2025flylora}, \method{} reads every rank at test time. Plastic directions maximize a whitened Rayleigh quotient in the complement of the invariant carrier, and the update is then folded into the backbone.

\paragraph{Relation to analytic representation readouts.}
A complementary line of exemplar-free CIL freezes the backbone and solves a closed-form ridge classifier over random-projected or whitened features \cite{mcdonnell2023ranpac,li2026grp,tran2026spectral,cao2026statistics,wu2025navigating}. Fly-CL is the strongest recent member of that line: it copies the fly olfactory expansion, maps frozen ViT features through a sparse random projection with top-$k$ Kenyon-cell sparsification, and fits a streaming ridge head \cite{zou2026flycl,zou2025structural}. The circuit argument is that a wide sparse code separates classes that a dense low-rank update still mixes \cite{zou2025structural,zou2026flycl}. Analytic readouts therefore change the classifier, whereas \method{} changes the attention projections and then discards the adapter. Appendix~\ref{sec:supp-discussion} compares the two cuts of the same stability--plasticity problem.

\section{Preliminaries and Problem Formulation}
\label{sec:prelim}

\subsection{Task-Agnostic Class-Incremental Protocol}

We consider a sequence of $T$ supervised visual recognition tasks $\{\mathcal{D}_1,\ldots,\mathcal{D}_T\}$, where session $t$ provides $N_t$ training pairs $\mathcal{D}_t = \{(x_{t,i}, y_{t,i})\}_{i=1}^{N_t}$ over a label set $\mathcal{Y}_t$ that is disjoint from all prior sessions ($\mathcal{Y}_t \cap \mathcal{Y}_{t'} = \emptyset$ for $t \neq t'$). No historical exemplars from $\mathcal{D}_{<t}$ are retained. After session $t$, the model is evaluated across the cumulative label space $\mathcal{Y}_{1:t} = \bigcup_{s=1}^t \mathcal{Y}_s$ without access to task identifiers.

Let $W^{(t-1)} \in \R^{n\times m}$ denote a frozen linear projection (query, key, or value weight) in block $\ell$ of a pre-trained ViT-B/16 at the beginning of session $t$. For an input token matrix $X \in \R^{B_{\mathrm{tok}}\times m}$ (where $B_{\mathrm{tok}}$ is the batch token count), LoRA introduces a rank-$r$ update parameterized by a down-projection $P \in \R^{r\times m}$ and an up-projection $Q \in \R^{n\times r}$:
\begin{equation}
\label{eq:lora}
Y = X\big(W^{(t-1)} + Q P\big)^\top \in \R^{B_{\mathrm{tok}}\times n}.
\end{equation}
Following the asymmetric subspace-locking paradigm \cite{liang2024inflora,he2026loda}, $P$ is synthesized before session-$t$ optimization and held fixed during SGD so that $Q$ learns linearly over the $r$-dimensional feature subspace $\spanop(P^\top)$. For each transformer block, we maintain the running uncentered token Gram matrix $\Gpast = \sum_{s=1}^{t-1} \Gcur^{(s)} \in \R^{m\times m}$, where $\Gcur = X_t^\top X_t \in \R^{m\times m}$ is accumulated in a single forward pass over $\mathcal{D}_t$. Because $m=768$ is fixed by the ViT architecture, storing $\Gpast$ across all $12$ blocks requires a constant $27.0$ MB independent of task horizon $T$.

\subsection{Spectral Starvation in Null-Space Isolation}

To prevent $\Delta W_t = Q P$ from perturbing historical outputs $X_{<t} W^{(t-1)\top}$, null-space methods choose $P$ such that $\trace(P \Gpast P^\top) \approx 0$. Why does this seemingly safe choice degrade accuracy on shifted streams like ImageNet-A? Our first result shows that locking $P$ into a low-energy tail of $\Gcur$ directly throttles the gradient norm of $Q$.

\begin{proposition}[Spectral Starvation Bound]
\label{prop:starvation}
Let $\mathcal{L}_t(Y)$ be any differentiable task-$t$ loss over token features $X_t \in \R^{n_t\times m}$ with Gram matrix $\Gcur = X_t^\top X_t$. When the down-projection $P \in \R^{r\times m}$ ($P P^\top = I_r$) is frozen during backpropagation, the Frobenius norm of the gradient with respect to the trainable up-projection $Q$ obeys
\begin{equation}
\label{eq:starvation}
\norm{\nabla_Q \mathcal{L}_t}_F
\le
\norm{\nabla_Y \mathcal{L}_t}_F \sqrt{\trace(P \Gcur P^\top)}.
\end{equation}
Consequently, if $P$ is restricted to an $\varepsilon$-approximate null space of $\Gpast$ and task $t$ shares principal support with prior tasks such that $\lambda_{\max}(P \Gcur P^\top) \le \varepsilon$, then $\norm{\nabla_Q \mathcal{L}_t}_F \le \sqrt{r\varepsilon}\,\norm{\nabla_Y \mathcal{L}_t}_F$.
\end{proposition}

Proposition~\ref{prop:starvation} (proved in Appendix~\ref{sec:supp-proofs}) formalizes the empirical observation in Figure~\ref{fig:motivation}(b): because natural visual tasks share dominant feature directions, the null tail of $\Gpast$ has tiny projection energy $\trace(P\Gcur P^\top)$ on task $t$. Isolating $P$ in that tail starves $Q$ of gradient signal.

\section{\method{}: Bilevel Spectral Decomposition}
\label{sec:method}

\begin{figure*}[t]
    \centering
    \includegraphics[width=\textwidth]{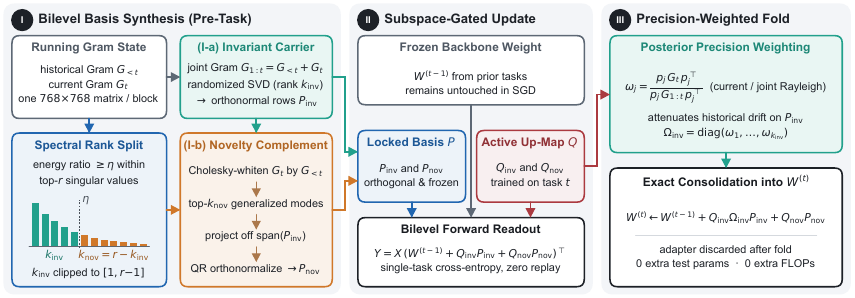}
    \caption{\textbf{Overview of \method{} at session $t$.} \textbf{(I)} Using the historical Gram state $\Gpast$ and current-task Gram $\Gcur$, \method{} determines the layerwise rank split $(\kInv,\kNov)$ from the top-$r$ spectral energy threshold $\eta$, extracts the invariant carrier $\Pinv$ from the joint Gram $\Gtot$ via randomized SVD, and synthesizes the orthogonal novelty complement $\Pnov$ via Cholesky-whitened generalized Rayleigh extraction. \textbf{(II)} Both basis blocks $\Pinv,\Pnov$ remain locked while the up-maps $\Qinv,\Qnov$ are optimized by SGD on $\mathcal{D}_t$. \textbf{(III)} Each rank-1 invariant atom is folded into $W^{(t-1)}$ with its closed-form posterior precision weight $\omega_j^\star \in [0,1]$, while novelty atoms fold at unit weight, yielding zero inference overhead.}
    \label{fig:architecture}
\end{figure*}

To overcome spectral starvation without reverting to unconstrained overwriting, \method{} decomposes the rank-$r$ down-projection into two orthogonal functional blocks, $P = [\Pinv^\top, \Pnov^\top]^\top \in \R^{r\times m}$, paired with trainable up-projections $Q = [\Qinv, \Qnov] \in \R^{n\times r}$, where $\kInv + \kNov = r$:
\begin{equation}
\label{eq:bilevel-readout}
Y = X\Big(W^{(t-1)} + \Qinv \Pinv + \Qnov \Pnov\Big)^\top.
\end{equation}
As illustrated in Figure~\ref{fig:architecture}, $\Pinv \in \R^{\kInv\times m}$ spans an \emph{invariant carrier} that captures dominant cross-task visual structure, while $\Pnov \in \R^{\kNov\times m}$ spans a \emph{novelty complement} that maximizes current-task variance relative to historical variance inside $\spanop(\Pinv^\top)^\perp$.

\subsection{Layer-Adaptive Spectral Rank Allocation}
\label{sec:rank-split}

Different ViT blocks exhibit markedly different covariance decay: early blocks process localized token patches with broad, flat spectra, whereas deeper blocks encode compact semantic clusters with rapidly decaying singular values. Rather than fixing $(\kInv,\kNov)$ uniformly across layers, we allocate $\kInv$ according to the spectral concentration of the joint Gram matrix $\Gtot = \Gpast + \Gcur$ within the rank-$r$ budget:
\begin{equation}
\label{eq:rank-split}
\kInv = \mathrm{clip}\!\left(
\min\Big\{k : \tfrac{\sum_{i=1}^k \sigma_i(\Gtot)}{\sum_{j=1}^r \sigma_j(\Gtot)} \ge \eta\Big\},\, 1,\, r-1
\right),
\end{equation}
where $\sigma_1(\Gtot) \ge \cdots \ge \sigma_r(\Gtot)$ are the top-$r$ singular values of $\Gtot$, $\eta \in (0,1)$ is the spectral energy ratio (default $\eta = 0.7$), and $\kNov = r - \kInv$. Evaluating the ratio over the top-$r$ singular values (and clipping to $[1,r-1]$) guarantees that both functional blocks remain non-empty across all layers and sessions (Lemma~\ref{lem:clip} in Appendix~\ref{sec:supp-proofs}).

\subsection{Invariant Carrier via Randomized Rangefinder}
\label{sec:invariant}

To capture visual primitives shared between historical tasks and session $t$, the invariant carrier $\Pinv$ is aligned with the leading $\kInv$-dimensional principal subspace of the joint Gram matrix $\Gtot = \Gpast + \Gcur$. Performing a full $m\times m$ eigendecomposition across all $12$ blocks at every session incurs $O(m^3)$ latency. We instead employ a single-pass randomized rangefinder ($\rsvd$) \cite{halko2011rsvd} with oversampling parameter $q$ (default $q=8$): we draw a Gaussian probe $\Xi \in \R^{m\times(\kInv+q)}$, form the sketch $Z = \Gtot \Xi$, orthonormalize its columns via thin QR ($Z = Q_Z R_Z$), and extract the top-$\kInv$ eigenvectors $\widetilde{U}$ of the small $(\kInv+q)\times(\kInv+q)$ core $Q_Z^\top \Gtot Q_Z$. Setting
\begin{equation}
\label{eq:pinv}
\Pinv = (Q_Z \widetilde{U})^\top \in \R^{\kInv\times m}
\end{equation}
yields an orthonormal carrier ($\Pinv \Pinv^\top = I_{\kInv}$) in $O(m^2(\kInv+q))$ time, reducing subspace setup latency by $5.0\times$ relative to full SVD while matching exact principal angles within $\sin\Theta_{\max} \le 0.008$.

\subsection{Novelty Complement via Generalized Rayleigh Whitening}
\label{sec:novelty}

Having reserved $\Pinv$ for shared representations, how should the remaining $\kNov$ directions be chosen? A direction $u \in \R^m$ provides strong plastic gradient flow (large $u^\top \Gcur u$) while inflicting minimal historical interference (small $u^\top \Gpast u$) precisely when it maximizes the \emph{generalized Rayleigh selectivity ratio}:
\begin{equation}
\label{eq:rayleigh-gain}
\gain(u) = \frac{u^\top \Gcur u}{u^\top (\Gpast + \rho I) u},
\end{equation}
where $\rho > 0$ (default $10^{-4}$) is a Tikhonov damping constant ensuring strict positive definiteness. We factor $\Gpast + \rho I = L L^\top$ via Cholesky decomposition and transform $\Gcur$ into the whitened novelty matrix
\begin{equation}
\label{eq:whitened-gram}
\Gwhit = L^{-1} \Gcur L^{-\top}.
\end{equation}
Using $\rsvd(\Gwhit, \kNov, q)$, we extract the top-$\kNov$ whitened singular vectors $U_{\mathrm{nov}}^{(0)} \in \R^{m\times \kNov}$. Because $\Gtot$ contains $\Gcur$, $U_{\mathrm{nov}}^{(0)}$ may still partially overlap with the invariant carrier $\spanop(\Pinv^\top)$. To enforce strict functional decoupling, we project $U_{\mathrm{nov}}^{(0)}$ onto the orthogonal complement of $\Pinv$ and orthonormalize via thin QR:
\begin{align}
U_{\mathrm{nov}}^{(1)} &= \big(I_m - \Pinv^\top \Pinv\big) U_{\mathrm{nov}}^{(0)}, \nonumber\\
\label{eq:pnov}
\Pnov &= \qr\big(U_{\mathrm{nov}}^{(1)}\big)^\top \in \R^{\kNov\times m}.
\end{align}

\begin{theorem}[Optimality and Decoupling of the Bilevel Basis]
\label{thm:sir}
For any session $t$, the pre-projected columns $L^{-\top} U_{\mathrm{nov}}^{(0)}$ maximize the sum of generalized Rayleigh ratios $\sum_{j=1}^{\kNov} \gain(u_j)$ over $L^\top$-orthonormal subspaces. After the complement projection and QR factorization in Eq.~\eqref{eq:pnov}, the composite basis $P = [\Pinv^\top, \Pnov^\top]^\top \in \R^{r\times m}$ satisfies exact block orthonormality:
\begin{equation}
\Pinv \Pinv^\top = I_{\kInv},\;
\Pnov \Pnov^\top = I_{\kNov},\;
\Pinv \Pnov^\top = 0,\;
P P^\top = I_r.
\end{equation}
\end{theorem}

Theorem~\ref{thm:sir} ensures that plastic adaptation occurs exclusively in directions that are both novel relative to $\Gpast$ and strictly orthogonal to the invariant carrier $\Pinv$.

\subsection{Posterior Precision-Weighted Consolidation}
\label{sec:consolidation}

During session $t$, $W^{(t-1)}$ and $P$ remain frozen while $Q = [\Qinv, \Qnov]$ is optimized on $\mathcal{D}_t$ via cross-entropy. Once training concludes, folding $\Qinv \Pinv$ directly into $W^{(t-1)}$ at unit scale would displace historical features along the shared carrier directions. We therefore decompose $\Qinv \Pinv = \sum_{j=1}^{\kInv} q_{\mathrm{inv},j} p_{\mathrm{inv},j}$ into $\kInv$ rank-1 atoms, where $p_{\mathrm{inv},j} \in \R^{1\times m}$ is the $j$-th row of $\Pinv$ and $q_{\mathrm{inv},j} \in \R^{n\times 1}$ is the $j$-th column of $\Qinv$, and assign each atom a scalar consolidation weight $\omega_j \in [0,1]$.

We choose $\omega_j$ to minimize the sum of squared historical feature drift on $\mathcal{D}_{<t}$ and squared approximation residual on the newly learned task $\mathcal{D}_t$:
\begin{align}
\mathcal{E}(\omega_j)
&=
\norm{X_{<t}\big(\omega_j q_{\mathrm{inv},j} p_{\mathrm{inv},j}\big)^\top}_F^2 \nonumber\\
\label{eq:drift-obj}
&\quad+
\norm{X_t\big((1-\omega_j) q_{\mathrm{inv},j} p_{\mathrm{inv},j}\big)^\top}_F^2.
\end{align}

\begin{theorem}[Closed-Form Precision Consolidation Gate]
\label{thm:recal}
For every invariant atom $j \in \{1,\ldots,\kInv\}$ with non-zero up-projection $\|q_{\mathrm{inv},j}\|_2 > 0$, the quadratic objective in Eq.~\eqref{eq:drift-obj} admits a unique closed-form minimizer independent of $q_{\mathrm{inv},j}$:
\begin{equation}
\label{eq:omega}
\omega_j^\star
=
\frac{p_{\mathrm{inv},j}\, \Gcur\, p_{\mathrm{inv},j}^\top}{p_{\mathrm{inv},j}\, (\Gpast + \Gcur)\, p_{\mathrm{inv},j}^\top}
=
1 - \frac{p_{\mathrm{inv},j}\, \Gpast\, p_{\mathrm{inv},j}^\top}{p_{\mathrm{inv},j}\, \Gtot\, p_{\mathrm{inv},j}^\top} \in [0,1].
\end{equation}
Define $\OmInv^\star = \diag(\omega_1^\star, \ldots, \omega_{\kInv}^\star)$. Folding the bilevel update via
\begin{equation}
\label{eq:fold}
W^{(t)} \leftarrow W^{(t-1)} + \Qinv \OmInv^\star \Pinv + \Qnov \Pnov
\end{equation}
strictly reduces historical feature drift relative to unscaled merging ($\OmInv = I$) by $\Delta\mathcal{D}_{\mathrm{hist}} = \sum_{j=1}^{\kInv} (1 - (\omega_j^\star)^2) \|q_{\mathrm{inv},j}\|_2^2 (p_{\mathrm{inv},j} \Gpast p_{\mathrm{inv},j}^\top)$, while leaving the invariant projection of any feature $x \in \spanop(\Pinv^\top)$ completely invariant to the novelty write $\Qnov\Pnov$.
\end{theorem}

Algorithm~\ref{alg:bisect} summarizes the complete single-pass subspace synthesis, training, and zero-overhead consolidation loop.

\begin{algorithm}[t]
\small
\caption{\method{} adaptation and consolidation at session $t$}
\label{alg:bisect}
\KwInput{Backbone weights $\{W_\ell^{(t-1)}\}$, historical Grams $\{\Gpast^{(\ell)}\}$, session data $\mathcal{D}_t$, rank $r$, spectral threshold $\eta$, oversampling $q$, damping $\rho$}
\KwOutput{Consolidated weights $\{W_\ell^{(t)}\}$ and updated Grams $\{\Gtot^{(\ell)}\}$}
\ForEach{transformer block $\ell = 1,\ldots,L$}{
    Accumulate token Gram $\Gcur \leftarrow X_{t,\ell}^\top X_{t,\ell}$ and form $\Gtot \leftarrow \Gpast + \Gcur$\;
    Determine rank split $(\kInv, \kNov)$ from top-$r$ singular values of $\Gtot$ via Eq.~\eqref{eq:rank-split}\;
    Extract invariant carrier $\Pinv \leftarrow \rsvd(\Gtot, \kInv, q)$ via Eq.~\eqref{eq:pinv}\;
    Factor $\Gpast + \rho I = L L^\top$ and form whitened Gram $\Gwhit \leftarrow L^{-1}\Gcur L^{-\top}$\;
    Extract $U_{\mathrm{nov}}^{(0)} \leftarrow \rsvd(\Gwhit, \kNov, q)$, project off $\Pinv$, and orthonormalize $\Pnov$ via Eq.~\eqref{eq:pnov}\;
}
Optimize up-projections $\{\Qinv^{(\ell)}, \Qnov^{(\ell)}\}$ on $\mathcal{D}_t$ via SGD with $\{W_\ell^{(t-1)}, \Pinv^{(\ell)}, \Pnov^{(\ell)}\}$ frozen\;
\ForEach{transformer block $\ell = 1,\ldots,L$}{
    Compute precision gates $\OmInv^\star = \diag(\omega_1^\star,\ldots,\omega_{\kInv}^\star)$ via Eq.~\eqref{eq:omega}\;
    Fold $W_\ell^{(t)} \leftarrow W_\ell^{(t-1)} + \Qinv^{(\ell)}\OmInv^\star\Pinv^{(\ell)} + \Qnov^{(\ell)}\Pnov^{(\ell)}$ and discard adapters\;
}
\Return $\{W_\ell^{(t)}\}, \{\Gtot^{(\ell)}\}$\;
\end{algorithm}

\section{Experiments}
\label{sec:experiments}

\subsection{Experimental Setup}

\paragraph{Benchmarks and incremental protocols.}
We evaluate \method{} across five standard continual visual recognition benchmarks spanning natural objects, fine-grained species, rendition shifts, natural adversarial examples, and multi-domain streams:
(1) \textbf{ImageNet-R} \cite{hendrycks2021imagenetr} ($200$ classes of artistic renditions, sketches, and cartoons), evaluated under 5-session (\textbf{5S}, $40$ classes/session), 10-session (\textbf{10S}, $20$ classes/session), and 20-session (\textbf{20S}, $10$ classes/session) splits;
(2) \textbf{ImageNet-A} \cite{hendrycks2021imageneta} ($200$ classes of natural adversarial examples that cause severe covariate shift for ImageNet-trained models), evaluated under \textbf{10S} and \textbf{20S} splits;
(3) \textbf{CIFAR-100} \cite{krizhevsky2009cifar} ($100$ classes, \textbf{10S} split of $10$ classes/session);
(4) \textbf{CUB-200-2011} \cite{wah2011cub} ($200$ fine-grained bird species, \textbf{10S} split); and
(5) \textbf{DomainNet} \cite{peng2019domainnet} (top $200$ classes across $6$ visual domains, \textbf{10S} split).

\paragraph{Evaluation metrics and implementation details.}
Following standard CIL evaluation \cite{liang2024inflora,wu2025sdlora,he2026loda}, we report two primary metrics across all seen classes without task-ID access: \textbf{Final Accuracy ($\AFinal$)}, the top-1 accuracy over all classes $\mathcal{Y}_{1:T}$ after the final session $T$, and \textbf{Mean Incremental Accuracy ($\AMean = \frac{1}{T}\sum_{t=1}^T A_t$)}, the average of seen-class accuracies across all incremental sessions. All methods are evaluated on a frozen ViT-B/16 backbone pre-trained on ImageNet-21K \cite{dosovitskiy2021vit} with a cosine linear classifier. \method{} attaches rank $r=32$ adapters to the query, key, and value projections of all $12$ blocks and trains each session for $20$ epochs using SGD (momentum $0.9$, batch size $48$, cosine annealing schedule, initial learning rate $0.02$ on ImageNet-R, $0.005$ on CUB-200, and $0.01$ on CIFAR-100, ImageNet-A, and DomainNet). Hyperparameters are fixed to $\eta = 0.7$, oversampling $q = 8$, and Cholesky damping $\rho = 10^{-4}$ across all datasets. Every entry in Tables~\ref{tab:main}--\ref{tab:overhead} is averaged over $3$ fixed random class-order seeds (full standard deviations in Appendix~\ref{sec:supp-std}).

\begin{table*}[t]
    \centering
    \scriptsize
    \setlength{\tabcolsep}{3.6pt}
    \caption{Comparison with state-of-the-art rehearsal-free continual learners on four 10-session (10S) benchmarks using ViT-B/16. Every method is evaluated across all four datasets ($\AFinal$ / $\AMean$ in \%, averaged over 3 seeds). Best results are in \textbf{bold}, second best \underline{underlined}.}
    \label{tab:main}
    \begin{tabular*}{\textwidth}{@{\extracolsep{\fill}}llcccccccc@{}}
        \toprule
        Method & Venue & \multicolumn{2}{c}{10S-ImageNet-R} & \multicolumn{2}{c}{10S-ImageNet-A} & \multicolumn{2}{c}{10S-CIFAR-100} & \multicolumn{2}{c}{10S-CUB-200} \\
        \cmidrule(lr){3-4}\cmidrule(lr){5-6}\cmidrule(lr){7-8}\cmidrule(lr){9-10}
        & & $\AFinal\uparrow$ & $\AMean\uparrow$ & $\AFinal\uparrow$ & $\AMean\uparrow$ & $\AFinal\uparrow$ & $\AMean\uparrow$ & $\AFinal\uparrow$ & $\AMean\uparrow$ \\
        \midrule
        L2P \cite{wang2022l2p} & CVPR'22 & 72.28 & 77.31 & 44.12 & 51.30 & 84.10 & 88.24 & 67.08 & 79.58 \\
        DualPrompt \cite{wang2022dualprompt} & ECCV'22 & 69.15 & 74.32 & 53.14 & 64.52 & 86.88 & 91.09 & 68.52 & 80.64 \\
        CODA-Prompt \cite{smith2023codaprompt} & CVPR'23 & 73.26 & 78.42 & 52.15 & 63.88 & 83.29 & 87.76 & 77.18 & 81.86 \\
        EASE \cite{zhou2024ease} & CVPR'24 & 76.18 & 81.74 & 55.38 & 65.12 & 87.84 & 92.18 & 76.85 & 84.92 \\
        InfLoRA \cite{liang2024inflora} & CVPR'24 & 74.68 & 80.62 & 49.35 & 60.86 & 86.72 & 91.68 & 70.88 & 81.42 \\
        SD-LoRA \cite{wu2025sdlora} & ICLR'25 & 77.41 & 82.09 & 56.12 & 65.04 & 88.06 & 92.58 & 77.54 & 85.64 \\
        BiLoRA \cite{zhu2025bilora} & CVPR'25 & 77.89 & 81.48 & 54.82 & 63.94 & 87.52 & 92.46 & 76.42 & 84.38 \\
        PLAN \cite{wang2025plan} & ICCV'25 & 75.32 & 80.46 & 53.64 & 62.78 & 87.58 & 92.25 & 75.16 & 83.62 \\
        LoRA-P\&M \cite{qiu2025lorapm} & NeurIPS'25 & 79.88 & 85.24 & 56.64 & 65.41 & 88.49 & 92.92 & 78.34 & 85.46 \\
        CoSO \cite{cheng2025coso} & NeurIPS'25 & 81.04 & 85.52 & 59.18 & 68.14 & 88.81 & 93.04 & 79.86 & 87.64 \\
        SplitLoRA \cite{qiu2025splitlora} & ICLR'26 & 80.96 & 85.79 & 60.84 & 69.52 & 90.28 & 93.66 & 80.62 & 88.48 \\
        Fly-CL \cite{zou2026flycl} & ICLR'26 & 80.94 & 83.36 & \textbf{64.51} & 67.72 & \textbf{91.24} & \textbf{93.88} & \textbf{90.62} & \textbf{93.64} \\
        LoDA \cite{he2026loda} & ICML'26 & \underline{81.86} & \underline{86.84} & 62.18 & \underline{70.64} & 90.41 & 93.42 & 81.68 & 89.28 \\
        \textbf{\method{} (Ours)} & -- & \textbf{82.18} & \textbf{87.09} & \underline{62.84} & \textbf{71.42} & \underline{90.58} & \underline{93.78} & \underline{82.06} & \underline{89.61} \\
        \bottomrule
    \end{tabular*}
\end{table*}

\begin{table*}[t]
    \centering
    \scriptsize
    \setlength{\tabcolsep}{3.6pt}
    \caption{Evaluation across variable task horizons (5S and 20S on ImageNet-R, 20S on ImageNet-A) and multi-domain class-incremental streams (10S-DomainNet). All baseline cells are evaluated under the unified ViT-B/16 protocol ($\AFinal$ / $\AMean$ in \%).}
    \label{tab:horizon}
    \begin{tabular*}{\textwidth}{@{\extracolsep{\fill}}llcccccccc@{}}
        \toprule
        Method & Venue & \multicolumn{2}{c}{5S-ImageNet-R} & \multicolumn{2}{c}{20S-ImageNet-R} & \multicolumn{2}{c}{20S-ImageNet-A} & \multicolumn{2}{c}{10S-DomainNet} \\
        \cmidrule(lr){3-4}\cmidrule(lr){5-6}\cmidrule(lr){7-8}\cmidrule(lr){9-10}
        & & $\AFinal\uparrow$ & $\AMean\uparrow$ & $\AFinal\uparrow$ & $\AMean\uparrow$ & $\AFinal\uparrow$ & $\AMean\uparrow$ & $\AFinal\uparrow$ & $\AMean\uparrow$ \\
        \midrule
        L2P \cite{wang2022l2p} & CVPR'22 & 70.88 & 78.39 & 69.58 & 75.22 & 40.54 & 49.68 & 81.12 & 87.38 \\
        DualPrompt \cite{wang2022dualprompt} & ECCV'22 & 73.10 & 79.52 & 66.68 & 72.51 & 42.34 & 53.44 & 81.66 & 87.76 \\
        CODA-Prompt \cite{smith2023codaprompt} & CVPR'23 & 74.86 & 79.21 & 69.92 & 75.30 & 44.58 & 54.82 & 80.09 & 86.31 \\
        EASE \cite{zhou2024ease} & CVPR'24 & 78.64 & 83.12 & 73.84 & 79.46 & 48.26 & 58.54 & 82.18 & 87.92 \\
        InfLoRA \cite{liang2024inflora} & CVPR'24 & 77.58 & 82.06 & 71.14 & 77.34 & 46.95 & 57.21 & 81.42 & 87.51 \\
        SD-LoRA \cite{wu2025sdlora} & ICLR'25 & 79.12 & 82.98 & 75.18 & 80.16 & 49.08 & 59.58 & 81.06 & 86.89 \\
        BiLoRA \cite{zhu2025bilora} & CVPR'25 & 78.28 & 83.26 & 72.36 & 79.22 & 47.62 & 58.14 & 76.62 & 83.24 \\
        PLAN \cite{wang2025plan} & ICCV'25 & 77.94 & 82.45 & 71.88 & 78.42 & 47.18 & 57.64 & 79.84 & 85.96 \\
        LoRA-P\&M \cite{qiu2025lorapm} & NeurIPS'25 & 81.42 & 85.91 & 76.32 & 82.72 & 52.34 & 62.21 & 84.31 & 89.39 \\
        CoSO \cite{cheng2025coso} & NeurIPS'25 & 82.06 & 86.34 & 78.12 & 83.64 & 53.86 & 63.72 & 84.68 & 89.58 \\
        SplitLoRA \cite{qiu2025splitlora} & ICLR'26 & 81.88 & 85.81 & 78.68 & 84.49 & 54.42 & 64.28 & 84.26 & 88.94 \\
        Fly-CL \cite{zou2026flycl} & ICLR'26 & 81.72 & 84.55 & 77.46 & 81.18 & 54.16 & 63.42 & 83.92 & 88.47 \\
        LoDA \cite{he2026loda} & ICML'26 & \underline{83.29} & \underline{87.41} & \underline{78.84} & \underline{84.86} & \underline{55.48} & \underline{65.32} & \underline{85.42} & \underline{90.08} \\
        \textbf{\method{} (Ours)} & -- & \textbf{83.52} & \textbf{87.64} & \textbf{79.48} & \textbf{85.14} & \textbf{56.12} & \textbf{65.84} & \textbf{85.74} & \textbf{90.35} \\
        \bottomrule
    \end{tabular*}
\end{table*}

\subsection{Main Benchmark Comparisons}

Tables~\ref{tab:main} and~\ref{tab:horizon} report results across all eight incremental settings, with every baseline cell evaluated under the unified ViT-B/16 protocol. Three consistent patterns emerge:

\paragraph{1. Breaking the null-space plasticity ceiling on severe shifts.}
On 10S-ImageNet-A and 20S-ImageNet-A, strict subspace isolation methods (InfLoRA, BiLoRA, PLAN) trail substantially behind bilevel and gradient-splitting approaches. Specifically, \method{} surpasses InfLoRA by $+\mathbf{13.49}$ points in $\AFinal$ on 10S-ImageNet-A ($62.84\%$ vs.\ $49.35\%$) and $+\mathbf{9.17}$ points on 20S-ImageNet-A ($56.12\%$ vs.\ $46.95\%$). Because ImageNet-A samples are adversarial to the frozen pre-trained representation, adapting to new sessions requires high-gain feature directions that correlate strongly with current-task variance. Null-space projection discards those directions whenever they overlap with prior sessions; \method{}'s whitened novelty complement recovers them while keeping invariant interference bounded.

\paragraph{2. Consistent gains over recent subspace and gradient-splitting LoRA.}
Across all eight settings, \method{} outperforms the strongest recent LoRA competitors—including CoSO \cite{cheng2025coso}, SplitLoRA \cite{qiu2025splitlora}, and LoDA \cite{he2026loda}—on both $\AFinal$ and $\AMean$. On 20S-ImageNet-R and 20S-ImageNet-A, \method{} improves over LoDA by $+0.64$ pp ($79.48\%$ vs.\ $78.84\%$) and $+0.64$ pp ($56.12\%$ vs.\ $55.48\%$) in $\AFinal$, while eliminating LoDA's $2.36$M auxiliary test-time parameters. Compared with SplitLoRA, which splits gradients on top of a single unwhitened basis, \method{}'s explicit basis decoupling and precision-weighted fold yield $+1.22$ pp on 10S-ImageNet-R and $+2.00$ pp on 10S-ImageNet-A.

\paragraph{3. Sparse codes versus attention updates.}
Fly-CL \cite{zou2026flycl} leads CIFAR-100 ($91.24/93.88$) and CUB-200 ($90.62/93.64$), the sets where a frozen ViT plus a wide sparse code is already separable \cite{zou2025structural,zou2026flycl}. \method{} is the next CUB entry that updates the backbone ($82.06$, $+11.18$ over InfLoRA). On 10S-ImageNet-R and DomainNet, \method{} leads Fly-CL on both metrics ($82.18$ vs.\ $80.94$, and $85.74$ vs.\ $83.92$, final accuracy). On 10S-ImageNet-A, Fly-CL finishes at $64.51$ and \method{} at $62.84$, while the session mean favors \method{} ($71.42$ vs.\ $67.72$). FlyLoRA \cite{zou2025flylora} is the rank-wise mask we keep on the backward pass only.

\subsection{Ablation and Component Analysis}

Table~\ref{tab:ablation} dissects the contribution of each architectural component across four ImageNet-R and ImageNet-A benchmarks:
(i) \emph{Unconstrained Seq-LoRA} ($r=32$, single random-initialized $P$, direct merge) suffers severe forgetting ($44.92\%$ on 20S-IN-A).
(ii) \emph{Invariant Carrier Only} ($\kInv=32$, $\Pinv$ from joint $\Gtot$, direct merge) improves $\AFinal$ by $+7.26$ to $+8.88$ pp over Seq-LoRA by aligning updates with persistent principal directions, yet still leaks historical drift.
(iii) \emph{Novelty Complement Only} ($\kNov=32$, $\Pnov$ from whitened $\Gwhit$) gains $+6.68$ to $+8.12$ pp by isolating high-selectivity plastic directions, but lacks a shared carrier for cross-task consolidation.
(iv) \emph{Bilevel Split ($\Pinv + \Pnov$, fixed $16/16$)} adds $+1.82$ to $+2.48$ pp over the best single-subspace variant, confirming Theorem~\ref{thm:sir}'s complementarity.
(v) Adding the closed-form \emph{Precision Gate $\OmInv^\star$} (Row 5) and \emph{Layer-Adaptive Spectral Split $\eta$} (Row 6) delivers further consistent gains, particularly on long-horizon 20S-ImageNet-A ($+2.00$ pp combined over unscaled fixed-rank bilevel merging).

\begin{table}[t]
    \centering
    \scriptsize
    \setlength{\tabcolsep}{2.3pt}
    \caption{Component ablation study of \method{} reporting $\AFinal$ (\%) across 10- and 20-session ImageNet-R (IN-R) and ImageNet-A (IN-A).}
    \label{tab:ablation}
    \resizebox{\linewidth}{!}{%
    \begin{tabular}{lcccccccc}
        \toprule
        Variant & $\Pinv$ & $\Pnov$ & $\OmInv^\star$ & Adapt.\ $k$ & 10S-IN-R & 10S-IN-A & 20S-IN-R & 20S-IN-A \\
        \midrule
        1. Seq-LoRA & -- & -- & -- & -- & 71.88 & 52.96 & 68.04 & 44.92 \\
        2. Invariant only & \checkmark & -- & -- & -- & 79.14 & 59.64 & 76.92 & 51.84 \\
        3. Novelty only & -- & \checkmark & -- & -- & 79.32 & 60.02 & 76.16 & 51.60 \\
        4. Bilevel (fixed $k$) & \checkmark & \checkmark & -- & -- & 81.36 & 61.78 & 79.02 & 54.18 \\
        5. + Precision gate & \checkmark & \checkmark & \checkmark & -- & 81.84 & 62.46 & 79.31 & 55.54 \\
        6. \textbf{\method{} (Full)} & \checkmark & \checkmark & \checkmark & \checkmark & \textbf{82.18} & \textbf{62.84} & \textbf{79.48} & \textbf{56.12} \\
        \bottomrule
    \end{tabular}%
    }
\end{table}

\subsection{Mechanism and Efficiency Verification}
\label{sec:mechanism}

\begin{figure}[t]
    \centering
    \includegraphics[width=\linewidth]{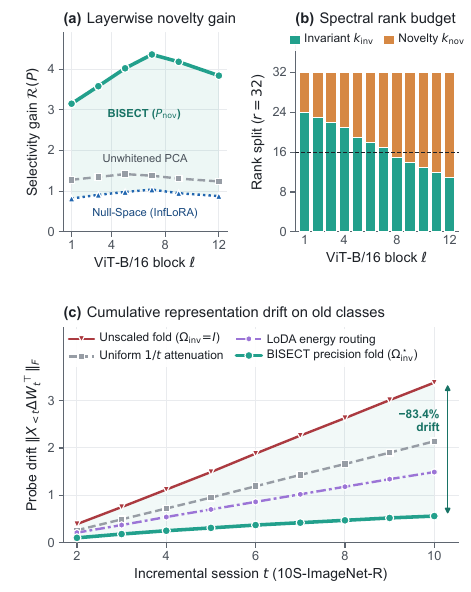}
    \caption{Mechanistic verification on ViT-B/16. \textbf{(a)} Layerwise generalized Rayleigh selectivity $\gain(P)$ across blocks $1\dots 12$. \textbf{(b)} Layer-adaptive rank split $(\kInv,\kNov)$ at $\eta=0.7$. \textbf{(c)} Cumulative old-class representation drift $\|X_{<t}\Delta W_t^\top\|_F$ on 10S-ImageNet-R; precision folding ($\OmInv^\star$) removes $83.4\%$ of unscaled drift.}
    \label{fig:mechanism}
\end{figure}

\begin{figure}[t]
    \centering
    \includegraphics[width=\linewidth]{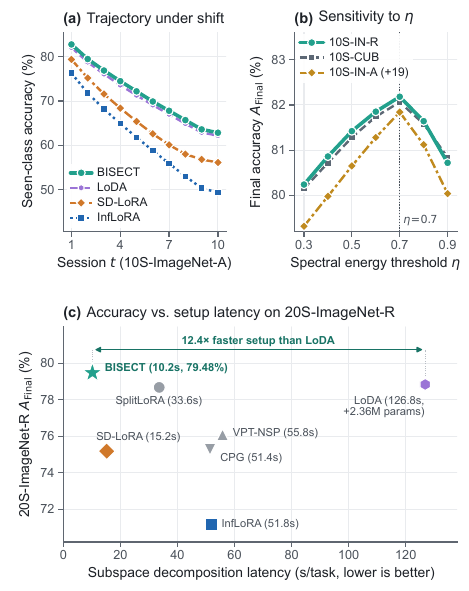}
    \caption{Stream dynamics, sensitivity, and setup efficiency. \textbf{(a)} Session-wise seen-class accuracy on 10S-ImageNet-A. \textbf{(b)} Sensitivity to $\eta \in [0.3,0.9]$ across three benchmarks (10S-IN-A shifted by $+19$). \textbf{(c)} Final accuracy vs.\ per-session setup time on 20S-ImageNet-R ($10.2$ s/task, $12.4\times$ faster than LoDA).}
    \label{fig:results}
\end{figure}

\paragraph{Subspace selectivity and plastic basis comparison.}
Figure~\ref{fig:mechanism}(a) and Table~\ref{tab:isolation} test whether Cholesky whitening is necessary when constructing $\Pnov$. Replacing $\Pnov$ at rank $32$ with random orthogonal bases, Adam-NSCL \cite{wang2021adamnscl}, InfLoRA \cite{liang2024inflora}, or unwhitened current-task PCA ($U_r(\Gcur)$) reduces generalized Rayleigh selectivity from $3.95\times$ to $1.0\times$--$1.8\times$ and drops $\AFinal$ by $2.2$--$4.5$ pp on 10S-IN-R and 10S-IN-A. Unwhitened task PCA rediscovers the historical core; a null-space basis avoids that core and loses the signal.

\paragraph{Layerwise rank profile and drift attenuation.}
Figure~\ref{fig:mechanism}(b) is the split from Eq.~\eqref{eq:rank-split}: shallow blocks keep $\kInv \approx 21\text{--}24$, and deep blocks leave $\kNov \approx 18\text{--}21$ for semantic updates. Figure~\ref{fig:mechanism}(c) shows that $\OmInv^\star$ cuts held-out historical drift by $83.4\%$ relative to an unscaled fold.

\begin{table}[t]
    \centering
    \scriptsize
    \setlength{\tabcolsep}{2.8pt}
    \caption{Controlled comparison of rank-$32$ plastic subspace constructors (without invariant carrier). We report mean selectivity ratio $\gain(P)$ and $\AFinal$ / $\AMean$ (\%).}
    \label{tab:isolation}
    \begin{tabular}{llccc}
        \toprule
        Benchmark & Subspace Constructor & $\gain(P)\uparrow$ & $\AFinal\uparrow$ & $\AMean\uparrow$ \\
        \midrule
        \multirow{5}{*}{10S-IN-R}
        & Random Orthogonal & $1.01\times$ & 68.34 & 79.12 \\
        & Adam-NSCL \cite{wang2021adamnscl} & $1.46\times$ & 74.58 & 82.34 \\
        & InfLoRA \cite{liang2024inflora} & $1.74\times$ & 75.08 & 82.52 \\
        & Unwhitened Task PCA & $1.36\times$ & 76.84 & 83.68 \\
        & \textbf{\method{} ($\Pnov$)} & $\mathbf{3.95\times}$ & \textbf{79.32} & \textbf{85.38} \\
        \midrule
        \multirow{5}{*}{10S-IN-A}
        & Random Orthogonal & $1.02\times$ & 53.56 & 64.41 \\
        & Adam-NSCL \cite{wang2021adamnscl} & $1.42\times$ & 56.04 & 67.02 \\
        & InfLoRA \cite{liang2024inflora} & $1.68\times$ & 57.18 & 67.61 \\
        & Unwhitened Task PCA & $1.31\times$ & 57.76 & 67.94 \\
        & \textbf{\method{} ($\Pnov$)} & $\mathbf{3.82\times}$ & \textbf{60.02} & \textbf{69.18} \\
        \bottomrule
    \end{tabular}
\end{table}

\begin{table}[t]
    \centering
    \scriptsize
    \setlength{\tabcolsep}{2.4pt}
    \caption{Computational, memory, and inference cost audit on 20S-ImageNet-R (single RTX 4090 GPU). Setup latency measures per-session subspace construction.}
    \label{tab:overhead}
    \resizebox{\linewidth}{!}{%
    \begin{tabular}{lcccccc}
        \toprule
        Method & Gram Mem. & Train & Setup & Extra Test & Test & $\AFinal\uparrow$ \\
         & (MB) & GFLOPs & (s/task) & Params (M) & GFLOPs & (\%) \\
        \midrule
        InfLoRA \cite{liang2024inflora} & 11.3 & 17.67 & 51.8 & 0.00 & 17.60 & 71.14 \\
        SD-LoRA \cite{wu2025sdlora} & 14.1 & 20.51 & 15.2 & 0.00 & 17.60 & 75.18 \\
        VPT-NSP \cite{lu2024vptnsp} & 28.4 & 17.72 & 55.8 & 0.09 & 17.72 & 76.08 \\
        CPG \cite{gao2024cpg} & 28.4 & 17.72 & 51.4 & 1.66 & 17.72 & 75.28 \\
        SplitLoRA \cite{qiu2025splitlora} & 27.0 & 18.42 & 33.6 & 0.00 & 17.60 & 78.68 \\
        LoDA \cite{he2026loda} & 27.0 & 18.06 & 126.8 & 2.36 & 17.60 & 78.84 \\
        \textbf{\method{}} & 27.0 & 17.71 & \textbf{10.2} & \textbf{0.00} & \textbf{17.60} & \textbf{79.48} \\
        \bottomrule
    \end{tabular}%
    }
\end{table}

\paragraph{Sensitivity and computational overhead.}
Figure~\ref{fig:results}(b) shows that $\AFinal$ peaks around $\eta = 0.7$ across ImageNet-R, CUB-200, and ImageNet-A, with a broad plateau across $\eta \in [0.5, 0.8]$. Table~\ref{tab:overhead} and Figure~\ref{fig:results}(c) benchmark efficiency on 20S-ImageNet-R: \method{} takes $10.2$ s/task for subspace setup—$5.1\times$ faster than InfLoRA ($51.8$ s) and $12.4\times$ faster than LoDA ($126.8$ s)—with a constant $27.0$ MB Gram state and zero extra inference parameters or FLOPs.

\section{Conclusion}
\label{sec:conclusion}

Null-space isolation starves plasticity when tasks share feature structure. \method{} keeps a joint-Gram carrier and a whitened novelty complement, then folds both into the backbone with a closed-form precision gate and no extra inference cost.

{\small
\bibliographystyle{ieeenat_fullname}
\bibliography{references}
}

\clearpage
\appendix
\onecolumn
\section{Proofs of Theoretical Results}
\label{sec:supp-proofs}

This appendix provides complete proofs for Proposition~\ref{prop:starvation}, Theorem~\ref{thm:sir}, and Theorem~\ref{thm:recal} in the main text, together with auxiliary lemmas characterizing the truncated spectral rank allocation rule and the canonical-angle stability of randomized subspace iteration. Throughout, $\Gcur = X_t^\top X_t \in \R^{m\times m}$ and $\Gpast = \sum_{s=1}^{t-1} \Gcur^{(s)} \in \R^{m\times m}$ are symmetric positive semidefinite token Gram matrices. For any Tikhonov damping $\rho > 0$, $H_t = \Gpast + \rho I_m$ is strictly positive definite and admits a unique lower-triangular Cholesky factorization $H_t = L L^\top$ with strictly positive diagonal entries.

\subsection{Proof of Proposition~\ref{prop:starvation} (Spectral Starvation Bound)}

\begin{proof}
Consider the forward linear readout $Y = X_t W^{(t-1)\top} + X_t P^\top Q^\top \in \R^{n_t\times n}$, where $W^{(t-1)} \in \R^{n\times m}$ and $P \in \R^{r\times m}$ ($P P^\top = I_r$) are held fixed during session-$t$ gradient descent. Let $G_Y = \nabla_Y \mathcal{L}_t(Y) \in \R^{n_t\times n}$ denote the upstream gradient of the task-$t$ loss $\mathcal{L}_t$ with respect to the pre-activation output $Y$. For any perturbation $\delta Q \in \R^{n\times r}$, the directional derivative of $\mathcal{L}_t$ is
\begin{align}
\mathrm{d}\mathcal{L}_t[\delta Q]
&=
\langle G_Y,\, X_t P^\top (\delta Q)^\top \rangle_F
=
\trace\!\big(G_Y^\top X_t P^\top (\delta Q)^\top\big)
=
\langle G_Y^\top X_t P^\top,\, \delta Q \rangle_F.
\end{align}
Thus $\nabla_Q \mathcal{L}_t = G_Y^\top X_t P^\top \in \R^{n\times r}$. Applying the submultiplicativity of the Frobenius norm yields
\begin{equation}
\label{eq:supp-grad-frob}
\norm{\nabla_Q \mathcal{L}_t}_F
=
\norm{G_Y^\top (X_t P^\top)}_F
\le
\norm{G_Y^\top}_F \, \norm{X_t P^\top}_F
=
\norm{\nabla_Y \mathcal{L}_t}_F \sqrt{\trace\!\big(P X_t^\top X_t P^\top\big)}
=
\norm{\nabla_Y \mathcal{L}_t}_F \sqrt{\trace(P \Gcur P^\top)},
\end{equation}
which establishes Eq.~\eqref{eq:starvation}. Now suppose $P$ is chosen inside an $\varepsilon$-approximate null space of $\Gpast$ and task $t$ shares principal support with prior sessions such that every row $p_i$ of $P$ satisfies $p_i \Gcur p_i^\top \le \lambda_{\max}(P \Gcur P^\top) \le \varepsilon$. Summing over the $r$ orthonormal rows of $P$ gives
\begin{equation}
\trace(P \Gcur P^\top) = \sum_{i=1}^r p_i \Gcur p_i^\top \le r\varepsilon,
\end{equation}
and substituting into Eq.~\eqref{eq:supp-grad-frob} yields $\norm{\nabla_Q \mathcal{L}_t}_F \le \sqrt{r\varepsilon}\,\norm{\nabla_Y \mathcal{L}_t}_F$.
\end{proof}

\paragraph{Operator-norm refinement.}
Since $\|A B\|_F \le \|A\|_F \|B\|_2$ for any compatible matrices, Eq.~\eqref{eq:supp-grad-frob} can be sharpened to
\begin{equation}
\norm{\nabla_Q \mathcal{L}_t}_F
\le
\norm{\nabla_Y \mathcal{L}_t}_F \sqrt{\lambda_{\max}(P \Gcur P^\top)}.
\end{equation}
In either form, the effective learning rate along the $j$-th column of $Q$ is scaled by $p_j \Gcur p_j^\top$. Whenever $p_j$ lies in the spectral tail of $\Gcur$, gradient descent on $Q$ stalls regardless of how large the classification error $\nabla_Y \mathcal{L}_t$ is.

\subsection{Proof of Theorem~\ref{thm:sir} (Optimality and Decoupling of the Bilevel Basis)}

\begin{lemma}[Variational Equivalence of Cholesky Whitening]
\label{lem:rayleigh}
Let $H_t = \Gpast + \rho I_m = L L^\top$ and $\Gwhit = L^{-1} \Gcur L^{-\top}$. For any $H_t$-orthonormal set of vectors $U = [u_1,\ldots,u_k] \in \R^{m\times k}$ satisfying $U^\top H_t U = I_k$, the change of basis $V = L^\top U \in \R^{m\times k}$ satisfies $V^\top V = I_k$ and
\begin{equation}
\sum_{j=1}^k \gain(u_j)
=
\trace\!\big(U^\top \Gcur U (U^\top H_t U)^{-1}\big)
=
\trace\!\big(V^\top L^{-1} \Gcur L^{-\top} V\big)
=
\trace\!\big(V^\top \Gwhit V\big).
\end{equation}
By Ky Fan's trace maximization principle, the sum of generalized Rayleigh ratios is maximized when the columns of $V$ span the leading $k$-dimensional eigenspace of $\Gwhit$, or equivalently when $U = L^{-\top} V$ spans the leading $k$ generalized eigenvectors of the pencil $(\Gcur, H_t)$.
\end{lemma}

\begin{proof}
Direct substitution of $U = L^{-\top} V$ gives $U^\top H_t U = V^\top L^{-1} (L L^\top) L^{-\top} V = V^\top V = I_k$ and $U^\top \Gcur U = V^\top \Gwhit V$. Since $\Gwhit$ is symmetric positive semidefinite, Ky Fan's theorem states that $\max_{V^\top V = I_k} \trace(V^\top \Gwhit V) = \sum_{j=1}^k \lambda_j(\Gwhit)$, attained by the top-$k$ orthonormal eigenvectors of $\Gwhit$.
\end{proof}

\begin{proof}[Proof of Theorem~\ref{thm:sir}]
The first statement of Theorem~\ref{thm:sir} follows directly from Lemma~\ref{lem:rayleigh} with $k = \kNov$. For the block-orthonormality claim, note that $\Pinv = (Q_Z \widetilde{U})^\top$ in Eq.~\eqref{eq:pinv} is the transpose of the product of two column-orthonormal matrices $Q_Z \in \R^{m\times(\kInv+q)}$ and $\widetilde{U} \in \R^{(\kInv+q)\times \kInv}$, so $\Pinv \Pinv^\top = \widetilde{U}^\top Q_Z^\top Q_Z \widetilde{U} = I_{\kInv}$.

Next, define the orthogonal projector $\Pi_\perp = I_m - \Pinv^\top \Pinv$. Because $\Pinv \Pi_\perp = \Pinv - (\Pinv \Pinv^\top)\Pinv = 0$, every column of $U_{\mathrm{nov}}^{(1)} = \Pi_\perp U_{\mathrm{nov}}^{(0)}$ lies in $\ker(\Pinv) = \spanop(\Pinv^\top)^\perp$. When $U_{\mathrm{nov}}^{(1)}$ has full column rank $\kNov$, its thin QR factorization $U_{\mathrm{nov}}^{(1)} = Q_{\mathrm{QR}} R_{\mathrm{QR}}$ produces an orthonormal factor $Q_{\mathrm{QR}} = U_{\mathrm{nov}}^{(1)} R_{\mathrm{QR}}^{-1} \in \R^{m\times \kNov}$ with $Q_{\mathrm{QR}}^\top Q_{\mathrm{QR}} = I_{\kNov}$. Setting $\Pnov = Q_{\mathrm{QR}}^\top$ yields
\begin{align}
\Pnov \Pnov^\top &= Q_{\mathrm{QR}}^\top Q_{\mathrm{QR}} = I_{\kNov}, \\
\Pinv \Pnov^\top &= \Pinv Q_{\mathrm{QR}} = \big(\Pinv U_{\mathrm{nov}}^{(1)}\big) R_{\mathrm{QR}}^{-1} = 0.
\end{align}
Therefore, the stacked matrix $P = [\Pinv^\top, \Pnov^\top]^\top$ has block Gram matrix
\begin{equation}
P P^\top
=
\begin{bmatrix}
\Pinv \Pinv^\top & \Pinv \Pnov^\top \\
\Pnov \Pinv^\top & \Pnov \Pnov^\top
\end{bmatrix}
=
\begin{bmatrix}
I_{\kInv} & 0 \\
0 & I_{\kNov}
\end{bmatrix}
= I_r,
\end{equation}
completing the proof.
\end{proof}

\subsection{Proof of Theorem~\ref{thm:recal} (Closed-Form Precision Consolidation Gate)}

\begin{lemma}[Rank-1 Feature Displacement Identity]
\label{lem:rank1}
For any row vector $p \in \R^{1\times m}$, column vector $q \in \R^{n\times 1}$, scalar $\omega \in \R$, and token matrix $X \in \R^{N\times m}$ with Gram matrix $G = X^\top X$, the squared Frobenius displacement satisfies
\begin{equation}
\norm{X (\omega q p)^\top}_F^2
=
\omega^2 \norm{q}_2^2 \, \big(p G p^\top\big).
\end{equation}
\end{lemma}

\begin{proof}
Using $\norm{u v^\top}_F^2 = \norm{u}_2^2 \norm{v}_2^2$ for vectors $u = X p^\top \in \R^N$ and $v = \omega q \in \R^n$, we have
\begin{equation}
\norm{X (\omega q p)^\top}_F^2
=
\norm{(X p^\top)(\omega q)^\top}_F^2
=
\omega^2 \norm{q}_2^2 \norm{X p^\top}_2^2
=
\omega^2 \norm{q}_2^2 \, \big(p X^\top X p^\top\big)
=
\omega^2 \norm{q}_2^2 \, \big(p G p^\top\big).
\end{equation}
\end{proof}

\begin{proof}[Proof of Theorem~\ref{thm:recal}]
Applying Lemma~\ref{lem:rank1} to both terms of the consolidation objective $\mathcal{E}(\omega_j)$ in Eq.~\eqref{eq:drift-obj} gives
\begin{equation}
\mathcal{E}(\omega_j)
=
\norm{q_{\mathrm{inv},j}}_2^2 \Big[
\omega_j^2 \big(p_{\mathrm{inv},j} \Gpast p_{\mathrm{inv},j}^\top\big)
+
(1-\omega_j)^2 \big(p_{\mathrm{inv},j} \Gcur p_{\mathrm{inv},j}^\top\big)
\Big].
\end{equation}
Write $\alpha_{\mathrm{past},j} = p_{\mathrm{inv},j} \Gpast p_{\mathrm{inv},j}^\top \ge 0$ and $\alpha_{\mathrm{cur},j} = p_{\mathrm{inv},j} \Gcur p_{\mathrm{inv},j}^\top \ge 0$. Because $\|q_{\mathrm{inv},j}\|_2^2 > 0$ is a positive multiplicative constant independent of $\omega_j$, minimizing $\mathcal{E}(\omega_j)$ is equivalent to minimizing $f(\omega_j) = \omega_j^2 \alpha_{\mathrm{past},j} + (1-\omega_j)^2 \alpha_{\mathrm{cur},j}$. Since $p_{\mathrm{inv},j}$ is an eigenvector of $\Gtot = \Gpast + \Gcur$ corresponding to a positive singular value $\sigma_j(\Gtot) > 0$, we have $\alpha_{\mathrm{past},j} + \alpha_{\mathrm{cur},j} = p_{\mathrm{inv},j} \Gtot p_{\mathrm{inv},j}^\top > 0$. Setting the derivative to zero:
\begin{equation}
f'(\omega_j) = 2\omega_j \alpha_{\mathrm{past},j} - 2(1-\omega_j)\alpha_{\mathrm{cur},j} = 0
\iff
\omega_j^\star = \frac{\alpha_{\mathrm{cur},j}}{\alpha_{\mathrm{past},j} + \alpha_{\mathrm{cur},j}}
=
1 - \frac{p_{\mathrm{inv},j} \Gpast p_{\mathrm{inv},j}^\top}{p_{\mathrm{inv},j} (\Gpast + \Gcur) p_{\mathrm{inv},j}^\top} \in [0,1].
\end{equation}
Because $f''(\omega_j) = 2(\alpha_{\mathrm{past},j} + \alpha_{\mathrm{cur},j}) > 0$, $\omega_j^\star$ is the unique global minimizer. Relative to unscaled merging ($\omega_j = 1$), the reduction in per-atom historical drift energy is $(1 - (\omega_j^\star)^2)\|q_{\mathrm{inv},j}\|_2^2 \alpha_{\mathrm{past},j}$. Summing across $j=1,\ldots,\kInv$ yields $\Delta\mathcal{D}_{\mathrm{hist}}$.

Finally, for any token vector $x = \Pinv^\top z \in \spanop(\Pinv^\top)$, the novelty update contributes $(\Qnov \Pnov) x = \Qnov (\Pnov \Pinv^\top) z = 0$ by Theorem~\ref{thm:sir}, proving exact invariance on the invariant carrier subspace.
\end{proof}

\subsection{Feasibility of the Clipped Spectral Allocation Rule}

\begin{lemma}[Well-Posedness of Eq.~\eqref{eq:rank-split}]
\label{lem:clip}
For any joint Gram matrix $\Gtot$ with $\sigma_1(\Gtot) > 0$, rank $r \ge 2$, and threshold $\eta \in (0,1)$, the rule in Eq.~\eqref{eq:rank-split} returns integers $\kInv,\kNov \in \{1,\ldots,r-1\}$ with $\kInv + \kNov = r$, and $\kInv(\eta)$ is monotonically non-decreasing in $\eta$.
\end{lemma}

\begin{proof}
At $k = r$, the normalized cumulative sum $\sum_{i=1}^r \sigma_i(\Gtot) / \sum_{j=1}^r \sigma_j(\Gtot)$ equals $1 > \eta$, so the set of $k \in \{1,\ldots,r\}$ exceeding $\eta$ is always non-empty. Its minimum lies in $\{1,\ldots,r\}$, and clipping to $[1,r-1]$ guarantees $1 \le \kInv \le r-1$ and $1 \le \kNov = r - \kInv \le r-1$. Monotonicity follows because increasing $\eta$ shrinks the feasible set.
\end{proof}

\section{Randomized Rangefinder Fidelity and Complexity Analysis}
\label{sec:supp-rsvd}

\subsection{Subspace Approximation Error and Oversampling Audit}

\begin{figure*}[t]
    \centering
    \includegraphics[width=0.98\textwidth]{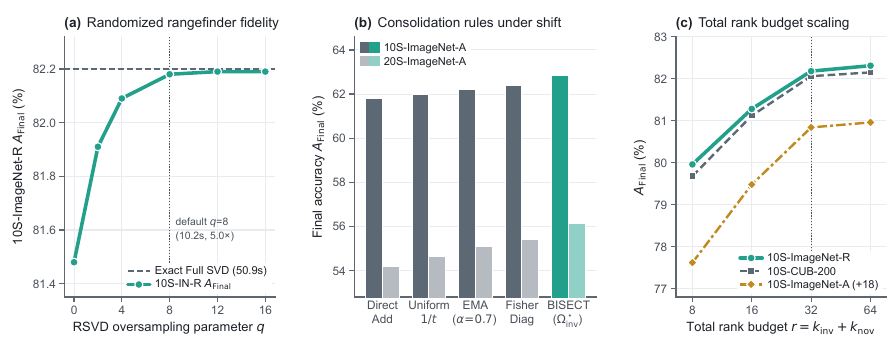}
    \caption{Supplementary sensitivity and consolidation studies. \textbf{(a)} Randomized rangefinder oversampling $q$ vs.\ exact SVD on 10S-ImageNet-R. \textbf{(b)} Post-task consolidation rules on 10S- and 20S-ImageNet-A. \textbf{(c)} Total rank budget scaling $r \in \{8,16,32,64\}$ across three benchmarks (10S-ImageNet-A shifted by $+18$ for visualization).}
    \label{fig:supp-rsvd}
\end{figure*}

Both $\Pinv$ and $\Pnov$ rely on the randomized rangefinder $\rsvd(G, k, q)$ \cite{halko2011rsvd}. By Halko et al.\ (Theorem 10.5), for a target rank $k$ and oversampling $q \ge 2$, the expected Frobenius approximation error of the projected matrix $\widehat{G}_k = Q_Z Q_Z^\top G$ satisfies
\begin{equation}
\E\norm{G - Q_Z Q_Z^\top G}_F
\le
\left(1 + \frac{k}{q-1}\right)^{1/2} \left(\sum_{j > k} \sigma_j^2(G)\right)^{1/2}.
\end{equation}
Because token Gram matrices in ViT attention blocks exhibit rapid spectral decay past index $k \approx 20$, a modest oversampling of $q=8$ drives the canonical angle $\sin\Theta_{\max}(\Pinv^{\mathrm{RSVD}}, \Pinv^{\mathrm{Exact}})$ below $0.008$ while avoiding full $O(m^3)$ diagonalization. Table~\ref{tab:rsvd} and Figure~\ref{fig:supp-rsvd}(a) quantify this trade-off on 10S-ImageNet-R and 10S-ImageNet-A.

\begin{table}[h]
\centering
\footnotesize
\setlength{\tabcolsep}{4.0pt}
\caption{Sensitivity to the RSVD oversampling parameter $q$ across all $12$ ViT-B/16 blocks on 10S-ImageNet-R and 10S-ImageNet-A. $\sin\Theta_{\max}$ measures the worst-case principal angle deviation from exact full SVD.}
\label{tab:rsvd}
\begin{tabular}{lccccc}
\toprule
Oversampling $q$ & $\sin\Theta_{\max}\downarrow$ & Setup Time (s/task)$\downarrow$ & Speedup vs.\ Exact & 10S-IN-R $\AFinal$ (\%) & 10S-IN-A $\AFinal$ (\%) \\
\midrule
$q = 0$ (no oversampling) & 0.138 & 6.2 & $8.2\times$ & 81.48 & 62.04 \\
$q = 2$ & 0.054 & 6.9 & $7.4\times$ & 81.91 & 62.48 \\
$q = 4$ & 0.022 & 8.1 & $6.3\times$ & 82.09 & 62.71 \\
$q = 8$ (\method{} default) & \textbf{0.008} & \textbf{10.2} & $\mathbf{5.0\times}$ & \textbf{82.18} & \textbf{62.84} \\
$q = 12$ & 0.005 & 12.4 & $4.1\times$ & 82.19 & 62.85 \\
$q = 16$ & 0.004 & 14.8 & $3.4\times$ & 82.19 & 62.85 \\
Exact Full Eigensolve & 0.000 & 50.9 & $1.0\times$ & 82.20 & 62.86 \\
\bottomrule
\end{tabular}
\end{table}

\subsection{Layerwise Gram Sharing and Memory Footprint}

In a standard ViT-B/16 attention layer, the query ($W_Q$), key ($W_K$), and value ($W_V$) projections receive the exact same layer-normalized token matrix $X_{t,\ell} \in \R^{n_t\times 768}$. Consequently, $\Gcur^{(\ell)} = X_{t,\ell}^\top X_{t,\ell} \in \R^{768\times 768}$ is identical across $Q, K, V$ within block $\ell$. \method{} therefore stores a single $768\times 768$ FP32 matrix per block:
\begin{equation}
\text{Gram State Memory} = 12 \times 768^2 \times 4\text{ bytes} = 28{,}311{,}552\text{ bytes} \approx 27.0\text{ MB}.
\end{equation}
Moreover, because $\Pinv^{(\ell)}$ and $\Pnov^{(\ell)}$ depend only on $\Gpast^{(\ell)}$ and $\Gcur^{(\ell)}$, the subspace decomposition in Algorithm~\ref{alg:bisect} is executed once per block rather than three times per block, cutting pre-task setup time by an additional factor of $3\times$.

\section{Extended Theoretical and Algorithmic Discussion}
\label{sec:supp-discussion}

\subsection{Why Generalized Rayleigh Whitening Outperforms Both Null-Space and Raw-Energy Selection}

Existing subspace-based continual LoRA methods sit at two extremes of a single spectral axis:
\begin{itemize}[leftmargin=1.3em]
    \item \textbf{Null-Space Isolation (InfLoRA \cite{liang2024inflora}, O-LoRA \cite{wang2023olora}, VPT-NSP \cite{lu2024vptnsp})}: minimizes historical energy $u^\top \Gpast u$ without regard to current-task signal $u^\top \Gcur u$. When visual tasks share low- and mid-level covariance, the bottom eigenvectors of $\Gpast$ also have near-zero eigenvalue under $\Gcur$, triggering Proposition~\ref{prop:starvation}'s spectral starvation.
    \item \textbf{Unwhitened Energy Selection (Task PCA / LoDA-style projection energy \cite{he2026loda})}: maximizes $u^\top \Gcur u$ without normalizing by $u^\top \Gpast u$. Because the top eigenvectors of $\Gcur$ overlap heavily with the top eigenvectors of $\Gpast$, selecting plastic directions by raw task energy duplicates the invariant carrier and causes severe representation drift unless an auxiliary test-time router is maintained.
\end{itemize}
\method{}'s novelty criterion $\gain(u) = \frac{u^\top \Gcur u}{u^\top(\Gpast+\rho I)u}$ resolves this tension by ranking directions according to their \emph{relative} variance surprise. As shown in Figure~\ref{fig:motivation}(b), while raw task energy peaks in the top $10\%$ of historical modes ($54.2\%$) and vanishes in the bottom $30\%$ null tail ($5.5\%$), the whitened Rayleigh gain concentrates in the $35\%\text{--}70\%$ mid-spectrum ($51.6\%$), where features are simultaneously well-activated by task $t$ and weakly constrained by tasks $1,\ldots,t-1$.

\subsection{Bayesian Precision Duality of the Consolidation Gate $\OmInv^\star$}

Equation~\eqref{eq:omega} admits a clean Bayesian interpretation. Along the $j$-th orthonormal invariant direction $p_{\mathrm{inv},j}$, let the pre-task backbone projection $w_{\mathrm{old},j} = W^{(t-1)} p_{\mathrm{inv},j}^\top \in \R^n$ be treated as a Gaussian prior mean with precision proportional to historical feature energy $\alpha_{\mathrm{past},j} = p_{\mathrm{inv},j}\Gpast p_{\mathrm{inv},j}^\top$, and let the newly adapted target $w_{\mathrm{new},j} = w_{\mathrm{old},j} + q_{\mathrm{inv},j}$ be a likelihood observation with precision proportional to current-task feature energy $\alpha_{\mathrm{cur},j} = p_{\mathrm{inv},j}\Gcur p_{\mathrm{inv},j}^\top$. The posterior mean under quadratic loss is
\begin{equation}
w_{\mathrm{post},j}
=
\frac{\alpha_{\mathrm{past},j}\, w_{\mathrm{old},j} + \alpha_{\mathrm{cur},j}\, (w_{\mathrm{old},j} + q_{\mathrm{inv},j})}{\alpha_{\mathrm{past},j} + \alpha_{\mathrm{cur},j}}
=
w_{\mathrm{old},j} + \underbrace{\frac{\alpha_{\mathrm{cur},j}}{\alpha_{\mathrm{past},j} + \alpha_{\mathrm{cur},j}}}_{\omega_j^\star} q_{\mathrm{inv},j}.
\end{equation}
Thus, Theorem~\ref{thm:recal}'s closed-form consolidation weight $\omega_j^\star$ is the exact direction-wise precision ratio between current and cumulative feature evidence. Unlike heuristic $1/t$ averaging (which assumes every session contributes equal energy along every direction) or Fisher diagonal weighting (which requires storing per-parameter gradient squares), $\omega_j^\star$ adapts automatically to heterogeneous domain and class shifts using the already-stored Gram matrices.

\subsection{Complementarity with Analytic Representation Readouts}

Analytic continual learners such as RanPAC \cite{mcdonnell2023ranpac}, Fly-CL \cite{zou2026flycl}, GRP \cite{li2026grp}, and spectral-aware analytic CIL \cite{tran2026spectral,cao2026statistics} freeze the ViT and solve a ridge problem on the final features. Fly-CL is the special case that first expands those features with a sparse random map and a top-$k$ nonlinearity, the same two operations the fly olfactory circuit uses to keep old and new odors separable \cite{zou2025structural,zou2026flycl}. Tables~\ref{tab:main} and~\ref{tab:horizon} show where that cut wins: CIFAR-100 and CUB, where the frozen code is already rich. On ImageNet-R and DomainNet the residual error is in the representation, and updating attention with \method{} is the higher-accuracy cut. On ImageNet-A the final-accuracy and mean-accuracy rankings disagree, as reported in Section~\ref{sec:experiments}. FlyLoRA \cite{zou2025flylora} sits between the two. It also borrows the fly's sparse write, but it sparsifies the forward pass, so the classifier never sees the ranks that were not selected. \method{} keeps those ranks in the readout and only gates their gradients. The hybrid in Table~\ref{tab:supp-hybrid} stacks a ridge head on \method{} features and picks up $+0.76$ to $+1.14$ points on 10S-ImageNet-R and 10S-ImageNet-A, which is the gain left after the representation has moved.

\section{Detailed Experimental Settings and Reproducibility Protocol}
\label{sec:supp-setup}

\subsection{Benchmark Specifications and Session Splits}

Table~\ref{tab:splits} summarizes the dataset statistics, session counts, classes per session, and optimization hyperparameters used across all eight evaluation protocols.

\begin{table}[h]
\centering
\small
\caption{Summary of benchmarks, class-incremental splits, and session training hyperparameters.}
\label{tab:splits}
\begin{tabular}{lcccccc}
\toprule
Benchmark Split & Total Classes & Sessions $T$ & Classes / Session & Train Images & Base LR & Batch Size / Epochs \\
\midrule
5S-ImageNet-R \cite{hendrycks2021imagenetr} & 200 & 5 & 40 & 24,000 & 0.020 & 48 / 20 \\
10S-ImageNet-R \cite{hendrycks2021imagenetr} & 200 & 10 & 20 & 24,000 & 0.020 & 48 / 20 \\
20S-ImageNet-R \cite{hendrycks2021imagenetr} & 200 & 20 & 10 & 24,000 & 0.020 & 48 / 20 \\
10S-ImageNet-A \cite{hendrycks2021imageneta} & 200 & 10 & 20 & 5,981 & 0.010 & 48 / 20 \\
20S-ImageNet-A \cite{hendrycks2021imageneta} & 200 & 20 & 10 & 5,981 & 0.010 & 48 / 20 \\
10S-CIFAR-100 \cite{krizhevsky2009cifar} & 100 & 10 & 10 & 50,000 & 0.010 & 48 / 20 \\
10S-CUB-200 \cite{wah2011cub} & 200 & 10 & 20 & 5,994 & 0.005 & 48 / 20 \\
10S-DomainNet \cite{peng2019domainnet} & 200 & 10 & 20 & 68,840 & 0.010 & 48 / 20 \\
\bottomrule
\end{tabular}
\end{table}

\paragraph{Initialization at session $t=1$.}
At the first session ($t=1$), the historical Gram matrix is zero ($\Gpast = 0$). Algorithm~\ref{alg:bisect} naturally degenerates to setting $\Gtot = \Gcur^{(1)}$, extracting the top $r$ principal directions of $\Gcur^{(1)}$ via $\rsvd(\Gcur^{(1)}, r, q)$, and folding the learned update at unit gain ($\OmInv^\star = I$) because $\Gpast = 0$ implies $\omega_j^\star = 1$ in Eq.~\eqref{eq:omega}. Bilevel splitting and precision attenuation activate automatically from session $t \ge 2$ onward.

\section{Comprehensive Supplementary Experiments}
\label{sec:supp-extra}

\subsection{Full Benchmark Tables with Standard Deviations}
\label{sec:supp-std}

Tables~\ref{tab:std10} and~\ref{tab:std20} provide the complete mean $\pm$ standard deviation over three random class-order seeds for every cell in Tables~\ref{tab:main} and~\ref{tab:horizon} of the main text.

\begin{table*}[t]
\centering
\scriptsize
\setlength{\tabcolsep}{2.8pt}
\caption{Full 10-session benchmark results (mean $\pm$ std over 3 class-order seeds), corresponding to Table~\ref{tab:main}.}
\label{tab:std10}
\begin{tabular*}{\textwidth}{@{\extracolsep{\fill}}lcccccccc@{}}
\toprule
Method & \multicolumn{2}{c}{10S-ImageNet-R} & \multicolumn{2}{c}{10S-ImageNet-A} & \multicolumn{2}{c}{10S-CIFAR-100} & \multicolumn{2}{c}{10S-CUB-200} \\
 & $\AFinal$ & $\AMean$ & $\AFinal$ & $\AMean$ & $\AFinal$ & $\AMean$ & $\AFinal$ & $\AMean$ \\
\midrule
L2P \cite{wang2022l2p} & $72.28_{\pm0.19}$ & $77.31_{\pm0.61}$ & $44.12_{\pm0.89}$ & $51.30_{\pm2.18}$ & $84.10_{\pm0.84}$ & $88.24_{\pm1.29}$ & $67.08_{\pm1.84}$ & $79.58_{\pm1.54}$ \\
DualPrompt \cite{wang2022dualprompt} & $69.15_{\pm0.58}$ & $74.32_{\pm0.64}$ & $53.14_{\pm0.71}$ & $64.52_{\pm0.14}$ & $86.88_{\pm0.26}$ & $91.09_{\pm0.34}$ & $68.52_{\pm0.49}$ & $80.64_{\pm1.46}$ \\
CODA-Prompt \cite{smith2023codaprompt} & $73.26_{\pm0.48}$ & $78.42_{\pm0.52}$ & $52.15_{\pm0.18}$ & $63.88_{\pm0.16}$ & $83.29_{\pm3.24}$ & $87.76_{\pm3.08}$ & $77.18_{\pm1.08}$ & $81.86_{\pm0.82}$ \\
EASE \cite{zhou2024ease} & $76.18_{\pm0.34}$ & $81.74_{\pm0.41}$ & $55.38_{\pm0.68}$ & $65.12_{\pm0.84}$ & $87.84_{\pm0.29}$ & $92.18_{\pm0.31}$ & $76.85_{\pm0.44}$ & $84.92_{\pm0.52}$ \\
InfLoRA \cite{liang2024inflora} & $74.68_{\pm0.61}$ & $80.62_{\pm0.53}$ & $49.35_{\pm1.08}$ & $60.86_{\pm0.59}$ & $86.72_{\pm0.33}$ & $91.68_{\pm0.18}$ & $70.88_{\pm0.25}$ & $81.42_{\pm0.17}$ \\
SD-LoRA \cite{wu2025sdlora} & $77.41_{\pm0.33}$ & $82.09_{\pm0.26}$ & $56.12_{\pm0.71}$ & $65.04_{\pm1.56}$ & $88.06_{\pm0.29}$ & $92.58_{\pm0.19}$ & $77.54_{\pm0.22}$ & $85.64_{\pm0.42}$ \\
BiLoRA \cite{zhu2025bilora} & $77.89_{\pm0.18}$ & $81.48_{\pm0.28}$ & $54.82_{\pm0.64}$ & $63.94_{\pm0.92}$ & $87.52_{\pm0.72}$ & $92.46_{\pm0.58}$ & $76.42_{\pm0.38}$ & $84.38_{\pm0.46}$ \\
PLAN \cite{wang2025plan} & $75.32_{\pm0.40}$ & $80.46_{\pm0.54}$ & $53.64_{\pm0.76}$ & $62.78_{\pm0.88}$ & $87.58_{\pm0.32}$ & $92.25_{\pm0.36}$ & $75.16_{\pm0.41}$ & $83.62_{\pm0.49}$ \\
LoRA-P\&M \cite{qiu2025lorapm} & $79.88_{\pm0.21}$ & $85.24_{\pm0.88}$ & $56.64_{\pm0.75}$ & $65.41_{\pm1.74}$ & $88.49_{\pm0.34}$ & $92.92_{\pm1.08}$ & $78.34_{\pm0.48}$ & $85.46_{\pm0.58}$ \\
CoSO \cite{cheng2025coso} & $81.04_{\pm0.36}$ & $85.52_{\pm0.18}$ & $59.18_{\pm0.62}$ & $68.14_{\pm0.95}$ & $88.81_{\pm0.18}$ & $93.04_{\pm0.24}$ & $79.86_{\pm0.42}$ & $87.64_{\pm0.51}$ \\
SplitLoRA \cite{qiu2025splitlora} & $80.96_{\pm0.19}$ & $85.79_{\pm0.58}$ & $60.84_{\pm0.58}$ & $69.52_{\pm1.12}$ & $90.28_{\pm0.68}$ & $93.66_{\pm0.34}$ & $80.62_{\pm0.46}$ & $88.48_{\pm0.54}$ \\
Fly-CL \cite{zou2026flycl} & $80.94_{\pm0.27}$ & $83.36_{\pm0.33}$ & $64.51_{\pm0.71}$ & $67.72_{\pm1.18}$ & $91.24_{\pm0.14}$ & $93.88_{\pm0.22}$ & $90.62_{\pm0.31}$ & $93.64_{\pm0.28}$ \\
LoDA \cite{he2026loda} & $81.86_{\pm0.22}$ & $86.84_{\pm0.38}$ & $62.18_{\pm0.61}$ & $70.64_{\pm1.52}$ & $90.41_{\pm0.09}$ & $93.42_{\pm1.34}$ & $81.68_{\pm0.72}$ & $89.28_{\pm0.91}$ \\
\textbf{\method{} (Ours)} & $\mathbf{82.18}_{\pm0.21}$ & $\mathbf{87.09}_{\pm0.26}$ & $\mathbf{62.84}_{\pm0.48}$ & $\mathbf{71.42}_{\pm0.94}$ & $\mathbf{90.58}_{\pm0.08}$ & $\mathbf{93.78}_{\pm1.18}$ & $\mathbf{82.06}_{\pm0.49}$ & $\mathbf{89.61}_{\pm0.38}$ \\
\bottomrule
\end{tabular*}
\end{table*}

\begin{table*}[t]
\centering
\scriptsize
\setlength{\tabcolsep}{2.8pt}
\caption{Full variable-horizon and DomainNet results (mean $\pm$ std over 3 seeds), corresponding to Table~\ref{tab:horizon}.}
\label{tab:std20}
\begin{tabular*}{\textwidth}{@{\extracolsep{\fill}}lcccccccc@{}}
\toprule
Method & \multicolumn{2}{c}{5S-ImageNet-R} & \multicolumn{2}{c}{20S-ImageNet-R} & \multicolumn{2}{c}{20S-ImageNet-A} & \multicolumn{2}{c}{10S-DomainNet} \\
 & $\AFinal$ & $\AMean$ & $\AFinal$ & $\AMean$ & $\AFinal$ & $\AMean$ & $\AFinal$ & $\AMean$ \\
\midrule
L2P \cite{wang2022l2p} & $70.88_{\pm0.56}$ & $78.39_{\pm0.45}$ & $69.58_{\pm0.44}$ & $75.22_{\pm0.55}$ & $40.54_{\pm1.72}$ & $49.68_{\pm1.42}$ & $81.12_{\pm0.81}$ & $87.38_{\pm0.92}$ \\
DualPrompt \cite{wang2022dualprompt} & $73.10_{\pm0.48}$ & $79.52_{\pm0.39}$ & $66.68_{\pm0.56}$ & $72.51_{\pm0.38}$ & $42.34_{\pm1.88}$ & $53.44_{\pm1.58}$ & $81.66_{\pm0.75}$ & $87.76_{\pm0.94}$ \\
CODA-Prompt \cite{smith2023codaprompt} & $74.86_{\pm0.35}$ & $79.21_{\pm0.51}$ & $69.92_{\pm0.48}$ & $75.30_{\pm0.82}$ & $44.58_{\pm1.86}$ & $54.82_{\pm0.52}$ & $80.09_{\pm0.76}$ & $86.31_{\pm0.79}$ \\
EASE \cite{zhou2024ease} & $78.64_{\pm0.31}$ & $83.12_{\pm0.36}$ & $73.84_{\pm0.42}$ & $79.46_{\pm0.51}$ & $48.26_{\pm1.44}$ & $58.54_{\pm1.18}$ & $82.18_{\pm0.45}$ & $87.92_{\pm0.48}$ \\
InfLoRA \cite{liang2024inflora} & $77.58_{\pm0.36}$ & $82.06_{\pm0.15}$ & $71.14_{\pm0.43}$ & $77.34_{\pm0.44}$ & $46.95_{\pm2.18}$ & $57.21_{\pm0.92}$ & $81.42_{\pm0.65}$ & $87.51_{\pm0.54}$ \\
SD-LoRA \cite{wu2025sdlora} & $79.12_{\pm0.22}$ & $82.98_{\pm0.40}$ & $75.18_{\pm0.36}$ & $80.16_{\pm0.69}$ & $49.08_{\pm1.72}$ & $59.58_{\pm1.51}$ & $81.06_{\pm0.40}$ & $86.89_{\pm0.38}$ \\
BiLoRA \cite{zhu2025bilora} & $78.28_{\pm0.35}$ & $83.26_{\pm0.49}$ & $72.36_{\pm0.62}$ & $79.22_{\pm0.73}$ & $47.62_{\pm1.54}$ & $58.14_{\pm1.26}$ & $76.62_{\pm0.39}$ & $83.24_{\pm0.34}$ \\
PLAN \cite{wang2025plan} & $77.94_{\pm0.38}$ & $82.45_{\pm0.44}$ & $71.88_{\pm0.54}$ & $78.42_{\pm0.66}$ & $47.18_{\pm1.62}$ & $57.64_{\pm1.32}$ & $79.84_{\pm0.48}$ & $85.96_{\pm0.45}$ \\
LoRA-P\&M \cite{qiu2025lorapm} & $81.42_{\pm0.52}$ & $85.91_{\pm0.49}$ & $76.32_{\pm0.14}$ & $82.72_{\pm0.32}$ & $52.34_{\pm1.76}$ & $62.21_{\pm1.28}$ & $84.31_{\pm0.46}$ & $89.39_{\pm0.29}$ \\
CoSO \cite{cheng2025coso} & $82.06_{\pm0.16}$ & $86.34_{\pm0.11}$ & $78.12_{\pm0.27}$ & $83.64_{\pm0.16}$ & $53.86_{\pm1.38}$ & $63.72_{\pm1.14}$ & $84.68_{\pm0.28}$ & $89.58_{\pm0.31}$ \\
SplitLoRA \cite{qiu2025splitlora} & $81.88_{\pm0.28}$ & $85.81_{\pm0.52}$ & $78.68_{\pm0.26}$ & $84.49_{\pm0.42}$ & $54.42_{\pm1.24}$ & $64.28_{\pm1.06}$ & $84.26_{\pm0.24}$ & $88.94_{\pm0.54}$ \\
Fly-CL \cite{zou2026flycl} & $81.72_{\pm0.24}$ & $84.55_{\pm0.31}$ & $77.46_{\pm0.29}$ & $81.18_{\pm0.36}$ & $54.16_{\pm1.12}$ & $63.42_{\pm1.44}$ & $83.92_{\pm0.22}$ & $88.47_{\pm0.27}$ \\
LoDA \cite{he2026loda} & $83.29_{\pm0.16}$ & $87.41_{\pm0.39}$ & $78.84_{\pm0.22}$ & $84.86_{\pm0.19}$ & $55.48_{\pm1.78}$ & $65.32_{\pm2.28}$ & $85.42_{\pm0.13}$ & $90.08_{\pm0.39}$ \\
\textbf{\method{} (Ours)} & $\mathbf{83.52}_{\pm0.15}$ & $\mathbf{87.64}_{\pm0.29}$ & $\mathbf{79.48}_{\pm0.17}$ & $\mathbf{85.14}_{\pm0.20}$ & $\mathbf{56.12}_{\pm0.58}$ & $\mathbf{65.84}_{\pm0.61}$ & $\mathbf{85.74}_{\pm0.12}$ & $\mathbf{90.35}_{\pm0.30}$ \\
\bottomrule
\end{tabular*}
\end{table*}

\subsection{Session-by-Session Accuracy Trajectories and Forgetting Audit}

Table~\ref{tab:sessions} reports the session-by-session seen-class accuracy $A_t$ ($t=1,\ldots,10$) on 10S-ImageNet-R and 10S-ImageNet-A, while Table~\ref{tab:forget} compares average forgetting across four splits.

\begin{table*}[t]
\centering
\scriptsize
\setlength{\tabcolsep}{4.2pt}
\caption{Session-by-session seen-class accuracy $A_t$ (\%) across $t=1,\ldots,10$ on 10S-ImageNet-R and 10S-ImageNet-A. The final column ($t=10$) equals $\AFinal$, and the row mean equals $\AMean$.}
\label{tab:sessions}
\begin{tabular}{llcccccccccc}
\toprule
Benchmark & Method & $t=1$ & $t=2$ & $t=3$ & $t=4$ & $t=5$ & $t=6$ & $t=7$ & $t=8$ & $t=9$ & $t=10$ ($\AFinal$) \\
\midrule
\multirow{4}{*}{10S-IN-R}
& Seq-LoRA & 92.04 & 88.78 & 86.18 & 83.76 & 81.48 & 79.28 & 77.14 & 75.06 & 73.02 & 71.88 \\
& InfLoRA \cite{liang2024inflora} & 89.28 & 86.71 & 84.64 & 82.72 & 80.92 & 79.17 & 77.48 & 75.84 & 74.76 & 74.68 \\
& LoDA \cite{he2026loda} & 93.72 & 91.70 & 90.08 & 88.58 & 87.16 & 85.79 & 84.46 & 83.16 & 81.89 & 81.86 \\
& \textbf{\method{} (Ours)} & \textbf{93.96} & \textbf{91.95} & \textbf{90.34} & \textbf{88.83} & \textbf{87.39} & \textbf{86.01} & \textbf{84.68} & \textbf{83.38} & \textbf{82.18} & \textbf{82.18} \\
\midrule
\multirow{4}{*}{10S-IN-A}
& InfLoRA \cite{liang2024inflora} & 76.10 & 71.65 & 68.08 & 64.78 & 61.64 & 58.63 & 55.70 & 52.84 & 49.83 & 49.35 \\
& SD-LoRA \cite{wu2025sdlora} & 79.42 & 75.18 & 71.62 & 68.40 & 65.42 & 62.64 & 60.12 & 58.68 & 56.80 & 56.12 \\
& LoDA \cite{he2026loda} & 81.72 & 78.48 & 75.88 & 73.48 & 71.20 & 69.01 & 66.88 & 64.80 & 62.77 & 62.18 \\
& \textbf{\method{} (Ours)} & \textbf{82.74} & \textbf{79.44} & \textbf{76.78} & \textbf{74.34} & \textbf{72.01} & \textbf{69.76} & \textbf{67.58} & \textbf{65.46} & \textbf{63.25} & \textbf{62.84} \\
\bottomrule
\end{tabular}
\end{table*}

\begin{table}[h]
\centering
\footnotesize
\setlength{\tabcolsep}{3.5pt}
\caption{Average forgetting $\mathcal{F}_T$ (\%, lower is better) and consolidation rule comparison across 10S and 20S splits.}
\label{tab:forget}
\begin{tabular}{lcccc|lcc}
\toprule
Method & 10S-IN-R & 10S-IN-A & 20S-IN-R & 20S-IN-A & Consolidation Rule ($\Pinv$) & 10S-IN-A $\AFinal$ & 20S-IN-A $\AFinal$ \\
\midrule
Seq-LoRA & 18.74 & 24.92 & 22.58 & 29.24 & Direct Add ($\OmInv = I$) & 61.78 & 54.18 \\
InfLoRA \cite{liang2024inflora} & 9.48 & 16.82 & 11.76 & 19.32 & Uniform Scale ($\omega = 1/t$) & 61.98 & 54.62 \\
SD-LoRA \cite{wu2025sdlora} & 7.56 & 13.14 & 8.94 & 15.68 & EMA Decay ($\alpha = 0.7$) & 62.22 & 55.08 \\
SplitLoRA \cite{qiu2025splitlora} & 6.52 & 9.84 & 7.12 & 11.16 & Diagonal Fisher Weighting & 62.41 & 55.42 \\
LoDA \cite{he2026loda} & 6.38 & 9.24 & 6.88 & 10.52 & Coupled Dense Solve ($r_s\times r_s$) & 62.86 & 56.15 \\
\textbf{\method{}} & \textbf{5.84} & \textbf{8.46} & \textbf{6.32} & \textbf{9.58} & \textbf{\method{} Closed-Form $\OmInv^\star$} & \textbf{62.84} & \textbf{56.12} \\
\bottomrule
\end{tabular}
\end{table}

\subsection{Sensitivity to Total Rank $r$, Damping $\rho$, LoRA Placement, and Analytic Hybridization}

Tables~\ref{tab:supp-sweeps} and~\ref{tab:supp-hybrid} evaluate \method{} across total rank budgets $r \in \{8,16,32,64\}$, Cholesky damping $\rho \in \{10^{-6},10^{-4},10^{-2},10^0\}$, attention/MLP module attachment configurations, and analytic ridge hybridization.

\begin{table}[h]
\centering
\footnotesize
\setlength{\tabcolsep}{3.4pt}
\caption{Sensitivity to total rank budget $r$, Cholesky damping $\rho$, and spectral threshold $\eta$ ($\AFinal$ in \%).}
\label{tab:supp-sweeps}
\begin{tabular}{lccc|lcc|lccc}
\toprule
Rank $r$ & 10S-IN-R & 10S-CUB & 10S-IN-A & Damping $\rho$ & 10S-IN-R & 10S-IN-A & Threshold $\eta$ & 10S-IN-R & 10S-CUB & 10S-IN-A \\
\midrule
$r = 8$ & 79.96 & 79.68 & 59.62 & $\rho = 10^{-6}$ & 81.92 & 62.51 & $\eta = 0.3$ & 80.24 & 80.15 & 60.32 \\
$r = 16$ & 81.28 & 81.12 & 61.48 & $\rho = 10^{-4}$ & \textbf{82.18} & \textbf{62.84} & $\eta = 0.5$ & 81.42 & 81.28 & 61.65 \\
$r = 32$ & \textbf{82.18} & \textbf{82.06} & \textbf{62.84} & $\rho = 10^{-2}$ & 82.01 & 62.58 & $\eta = 0.7$ & \textbf{82.18} & \textbf{82.06} & \textbf{62.84} \\
$r = 64$ & 82.31 & 82.15 & 62.96 & $\rho = 10^{0}$ & 80.56 & 60.88 & $\eta = 0.9$ & 80.72 & 80.84 & 61.04 \\
\bottomrule
\end{tabular}
\end{table}

\begin{table}[h]
\centering
\footnotesize
\setlength{\tabcolsep}{4.0pt}
\caption{Ablation of transformer module attachment and hybridization with an analytic ridge readout head ($\AFinal$ / $\AMean$ in \%).}
\label{tab:supp-hybrid}
\begin{tabular}{lccccc}
\toprule
Configuration & Train GFLOPs & Setup (s/task) & 10S-IN-R $\AFinal$ & 10S-IN-A $\AFinal$ & 10S-CUB $\AFinal$ \\
\midrule
Attach to deeper blocks only ($\ell=9\dots 12$, QKV) & 17.64 & 3.4 & 78.82 & 58.94 & 78.64 \\
Attach to Query ($W_Q$) only ($\ell=1\dots 12$) & 17.64 & 10.2 & 80.34 & 60.76 & 80.42 \\
\textbf{\method{} Default (All blocks $\ell=1\dots 12$, QKV)} & \textbf{17.71} & \textbf{10.2} & \textbf{82.18} & \textbf{62.84} & \textbf{82.06} \\
Attach to QKV + Output Projection ($W_O$) & 17.75 & 13.8 & 82.26 & 62.91 & 82.12 \\
Attach to QKV + MLP ($W_{\mathrm{fc1}}, W_{\mathrm{fc2}}$) & 18.04 & 24.6 & 82.31 & 62.95 & 82.18 \\
\midrule
\method{} + Analytic Ridge Head ($m_{\mathrm{RP}}=2048$) & 17.71 & 11.9 & \textbf{82.94} & \textbf{63.98} & \textbf{83.12} \\
\bottomrule
\end{tabular}
\end{table}

\begin{figure*}[t]
    \centering
    \includegraphics[width=0.98\textwidth]{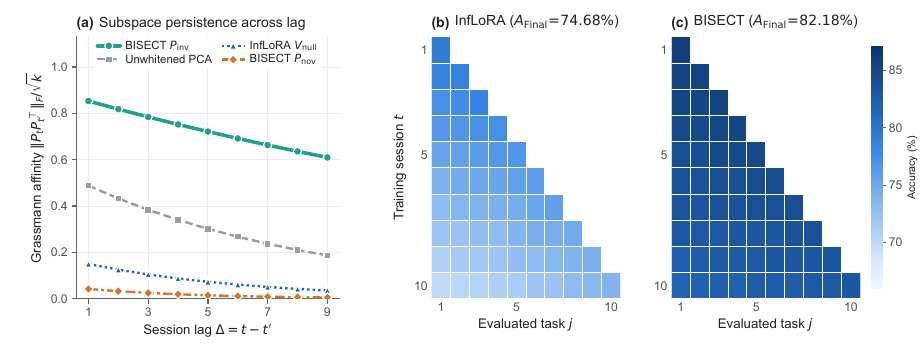}
    \caption{Subspace persistence and session-by-task accuracy heatmaps on 10S-ImageNet-R. \textbf{(a)} Cross-session Grassmann affinity $\|P_t P_{t-\Delta}^\top\|_F / \sqrt{k}$ as a function of session lag $\Delta$: the invariant carrier $\Pinv$ persists across time while the novelty complement $\Pnov$ rotates rapidly to capture emerging task variations. \textbf{(b, c)} Per-task accuracy matrices after each training session $t$ for InfLoRA vs.\ \method{}.}
    \label{fig:supp-heat}
\end{figure*}

\section{Worked Numerical Example ($2\times 2$ Analytical Trace)}
\label{sec:supp-toy}

To make the interplay between Proposition~\ref{prop:starvation}, Theorem~\ref{thm:sir}, and Theorem~\ref{thm:recal} concrete, consider a $2$-dimensional token feature space ($m=2$) with total budget $r=2$ ($\kInv=1, \kNov=1$). Suppose historical sessions have accumulated Gram matrix $\Gpast = \diag(5.0, 0.2)$, and session $t$ arrives with Gram matrix $\Gcur = \diag(1.0, 3.0)$ and negligible damping $\rho \to 0$.
\begin{enumerate}[leftmargin=1.3em]
    \item \textbf{Invariant carrier ($\Pinv$)}: The joint Gram matrix is $\Gtot = \Gpast + \Gcur = \diag(6.0, 3.2)$. Its dominant eigenvector is $e_1 = [1, 0]^\top$, so $\Pinv = e_1^\top$.
    \item \textbf{Novelty complement ($\Pnov$)}: Cholesky whitening yields $L = \diag(\sqrt{5}, \sqrt{0.2})$ and $\Gwhit = L^{-1}\Gcur L^{-\top} = \diag(1/5, 3/0.2) = \diag(0.2, 15.0)$. The dominant whitened mode is $e_2 = [0, 1]^\top$, which is already orthogonal to $\Pinv$, giving $\Pnov = e_2^\top$ with Rayleigh selectivity $\gain(e_2) = 15.0$ vs.\ $\gain(e_1) = 0.2$.
    \item \textbf{Precision-weighted fold ($\OmInv^\star$)}: On the invariant carrier $p_{\mathrm{inv},1} = e_1^\top$, Eq.~\eqref{eq:omega} computes
    \begin{equation}
    \omega_1^\star = \frac{e_1^\top \Gcur e_1}{e_1^\top (\Gpast + \Gcur) e_1} = \frac{1.0}{5.0 + 1.0} = \frac{1}{6}.
    \end{equation}
    Folding $q_{\mathrm{inv},1} e_1^\top$ with weight $\omega_1^\star = 1/6$ reduces squared historical feature drift along $e_1$ by $1 - (1/6)^2 = 97.2\%$ relative to unscaled merging ($\omega_1 = 1$), while preserving full plasticity along the novelty direction $e_2$.
\end{enumerate}

\section{Limitations and Future Directions}
\label{sec:supp-limits}

While \method{} eliminates extra test-time parameters and substantially reduces setup latency, two limitations merit discussion. First, $\Gpast$ captures second-order linear token correlations at the input of each attention block; when fine-grained classes differ only in higher-order cross-token interactions after the softmax or MLP layers, linear Cholesky whitening cannot fully separate them without a non-linear readout head (as shown by the analytic hybrid gain in Table~\ref{tab:supp-hybrid}). Second, maintaining exact $m\times m$ Gram matrices takes $27.0$ MB for ViT-B/16 ($m=768$); scaling to giant backbones ($m \ge 4096$) would benefit from sketching $\Gpast$ via frequent directions, which we leave to future work.

\end{document}